\documentclass[lettersize,journal]{IEEEtran}
\usepackage{amsmath,amsfonts}
\usepackage{bm}
\usepackage{hyperref}
\usepackage{algorithmic}
\usepackage{algorithm}
\usepackage{array}
\usepackage[caption=false,font=normalsize,labelfont=sf,textfont=sf]{subfig}
\usepackage{textcomp}
\usepackage{stfloats}
\usepackage{url}
\usepackage{verbatim}
\usepackage{graphicx}
\usepackage{cite}
\usepackage{booktabs}
\usepackage{multirow}
\usepackage{xcolor}
\newcommand{\etal}{\textit{et al.}}

\newcommand{\ie}{\textit{i.e.}}
\usepackage{amssymb}
\begin{document}

\title{Breaking Spatial Uniformity: Prior-Guided Mamba with \\Radial Serialization for Lens Flare Removal}

\author{Zijia~Fu,
        Yuanfei~Huang,~\IEEEmembership{Member,~IEEE,}
        Lizhi~Wang,~\IEEEmembership{Member,~IEEE,}
        and~Hua~Huang,~\IEEEmembership{Senior Member,~IEEE}%
\thanks{The authors are with the School of Artificial Intelligence, Beijing Normal University, Beijing 100875, China. Email: purplehomee@icloud.com, \{yfhuang, lizhiwang, huahuang\}@bnu.edu.cn. \\(\textit{Corresponding author: Yuanfei Huang})} 
}

\maketitle

\begin{abstract}
Lens flares, caused by complex optical aberrations, severely degrade image quality especially in nighttime photography. Although recent restoration methods have made remarkable progress, most still rely on spatially uniform processing. They are failing to handle the region-dependent restoration demands of flare scenes, where saturated light sources should be preserved, flare artifacts removed, and background details recovered. To address this challenge, we propose DeflareMambav2, a prior-guided Mamba framework for lens flare removal. Specifically, we introduce a Flare Prior Network (FPN) to estimate flare priors and guide adaptive restoration. Besides, a novel radial serialization strategy breaks spatially homogeneous processing by performing flare-aware targeted sampling, and better supports long-range modeling in State Space Models (SSMs). Based on these priors, the backbone adopts a dual-level adaptive scheme. It explicitly preserves light-source regions to avoid over-processing, and applies curriculum-based restoration to the remaining contaminated areas while calibrating restoration intensity at the pixel level. Extensive experiments demonstrate that DeflareMambav2 achieves state-of-the-art performance with reduced parameter burden. Code is available in \href{https://github.com/BNU-ERC-ITEA/DeflareMambav2}{https://github.com/BNU-ERC-ITEA/DeflareMambav2}.


\end{abstract}


\section{INTRODUCTION}
\IEEEPARstart{L}{ens} flare is a prevalent optical artifact encountered in images captured under an intense optical source. As complex artificial lighting becomes ubiquitous in nighttime environments, these artifacts are significantly amplified, severely degrading imaging quality and obscuring critical details. Consequently, effectively disentangling such artifacts from intrinsic scene information is crucial.


Flare removal aims to suppress flare artifacts while preserving valid light sources. However, existing deep learning methods struggle with the conflicting restoration goals in flare. Wu~\etal~\cite{WuY2021ICCV} first pointed out that the main difficulty is to remove the flare without damaging the light source. This remains challenging even with semi-synthetic data, as separating the light source from the flare is not straightforward. To handle this problem, they used a threshold-based mask to refine the output. Yet this heuristic strategy often fails on highly saturated glare streaks. It may preserve bright flare regions as valid light sources. Dai~\etal~\cite{DaiY2022NeurIPS,DaiY2024TPAMI} further introduced the Flare7K++ synthetic pipeline. Their method employs dual supervision, asking the model to predict both the flare image and the clean light-preserved image. This design encourages the network to separate flare artifacts from valid scene content. However, it still depends on strong assumptions about region separation. It may become less effective when flare structures are complex or mixed with background details.

\begin{figure}[!t]
\centering
\includegraphics[width=3.5in]{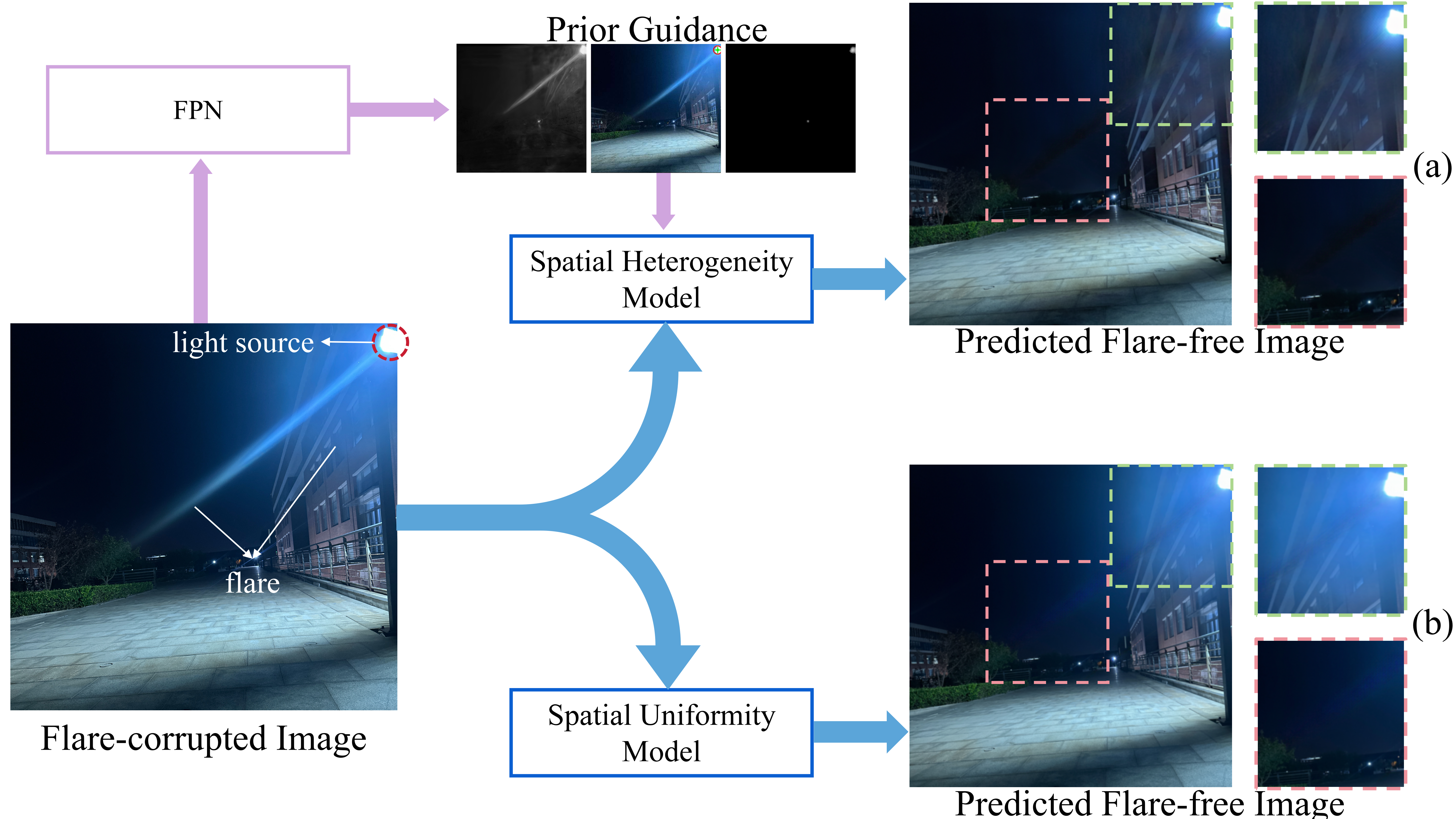}
\vspace{-0.2cm}
\caption{Comparison of lens flare removal strategies: (a) Output from our DeflareMambav2, which leverages prior to enable spatially heterogeneous processing; (b) Output from Uformer, which is a typical spatially uniform processing strategy using pre-trained weights from Flare7K++~\cite{DaiY2024TPAMI}.}
\label{fig1}
\end{figure}

Despite the effectiveness of this data-driven supervision, a critical architectural bottleneck remains. Current architectures typically employ spatially uniform processing across image. Both CNN-based and Transformer-based models suffer from this limitation. CNN methods~\cite{WuY2021ICCV,Sharma2021CVPR,Liangyu2021CVPRW,Zamir2021CVPR} rely on spatially invariant kernels, which struggle to simultaneously suppress flare while preserving light sources when both components share nearly identical high-intensity features. Similarly, Transformer models~\cite{DaiY2022NeurIPS,DengH2024TIP,DaiY2024TPAMI,Kotp2024ICASSP,ChenG2025TCSVT,JieZ2025ICCV,MaT2025TASE} compute self-attention based on token similarity. The high correlation between flare and light sources causes the attention mechanism to aggregate their features indistinguishably. Subsequently, DeflareMamba~\cite{HuangY2025ACMMM} made significant progress by adopting hierarchical scanning strategies to capture global and local contextual dependencies. It established long-range pixel correlations through varied-stride sampling within State Space Models, effectively mitigating the limitations of pure CNN or Transformer methods. However, DeflareMamba still relies on spatially uniform serialization of image regions. Flare artifacts requiring removal and light sources requiring preservation are indiscriminately encoded into the same hidden states without spatial awareness. Such undifferentiated processing risks contaminating the SSM's hidden representations with contradictory information. As illustrated in Fig.~\ref{fig1}, {\em to overcome this bottleneck, it is urgent to move beyond uniform serialization toward a strategy that respects the underlying geometric properties of flare phenomena}.


To address this limitation, we leverage optical priors from lens flare physics. Lens flares can be categorized into two types based on their physical origins: reflective flare and scattering flare. For reflective flare, Koreban~\etal~\cite{KorebanF2009ICCP} identified the optical center symmetry prior, noting that flare artifacts and primary light sources exhibit point symmetry relative to the lens optical center. However, existing methods overlook the scattering flare component, which dominates in night photography and exhibits fundamentally different spatial characteristics. Unlike reflective flare's center-symmetric pattern, scattering flare exhibits radial decay properties. Intensity diminishes with spatial distance from the light source in a radially symmetric manner, which has been largely ignored in current methods. {\em Understanding these different geometric properties is essential for designing serialization strategies that can differentially process conflicting restoration objectives}. 
To break this bottleneck and harness the scattering flare prior, we propose DeflareMambav2. It resolves the intrinsic conflict between flare removal and light source preservation, and leverages physical prior to model local-global feature dependencies more effectively than DeflareMamba~\cite{HuangY2025ACMMM}. Main contributions are summarized as follows:
\begin{itemize}
    \item We propose DeflareMambav2 via decoupling the processing of the light source from flare-corrupted areas. It overcomes the performance bottlenecks inherent to traditional spatially uniform processing strategies.

    \item We introduce a multi-illuminant training paradigm for the prior network. It extracts high-fidelity physical priors in complex multi-light scenarios. Guided by these precise priors, the DeflareMambav2 network effectively mitigates performance degradation in multi-source flare restoration.

    \item We establish a prior-driven restoration mechanism. Three physical priors extracted by the Flare Prior Network are translated into three customized functional modules. This integration provides precise physics-aware guidance throughout the feature modulation process.
\end{itemize}





\section{RELATED WORK}
\subsection{Lens Flare Removal}
Early traditional methods generally fall into two categories: physical solutions and computational approaches. The former, employing specialized lenses, hoods, and anti-reflective coatings, often incur high hardware costs and struggle to maintain effectiveness under severe glare contamination. Alternatively, early computational methods \cite{GaborK2013SPIE,CSAshaICDP2019,KorebanF2009ICCP} commonly lack robustness in complex illumination environments.
Building upon the foundation laid by the Flare7K++ dataset~\cite{DaiY2024TPAMI} and its training pipeline, recent literature has witnessed remarkable advancements in flare removal. Initially, Kotp and Torki \cite{Kotp2024ICASSP} proposed a novel framework that integrates a depth estimation module into the Uformer \cite{WangZ2022CVPR} architecture, demonstrating promising restoration capabilities. DeflareMamba \cite{HuangY2025ACMMM} introduced the efficient sequence modeling capabilities of State Space Models (SSMs) to capture the distinct dependencies of local scattering and global reflective flares. Furthermore, Chen \etal \cite{ChenG2025TCSVT} proposed the LPFSformer, which incorporates location prior guidance and a global hybrid feature compensation mechanism, providing key inspiration for this work. From a frequency-domain perspective, Dong \etal \cite{DongW2026TCSVT} introduced SAFAformer, integrating frequency-adaptive guidance with scale-aware Transformer modeling. To further address complex non-uniform scattered flares, Zhu \etal \cite{JieZ2025ICCV} recently introduced FGRNet, which exploits spatial-frequency features for holistic local and global feature extraction. Ma \etal \cite{MaT2025TASE} proposed SGSFT, which extracts intrinsic scene priors to semantically decouple flares from light sources, effectively guiding flare removal in complex real-world scenes.



\subsection{Degradation-Aware Priors}
Using specific priors to guide deep networks is a highly effective strategy in image restoration. Previous works have successfully applied various priors to tasks like denoising~\cite{HuangY2025IJCV,DaH2026TMM}, deraining~\cite{ZhangR2024TMM}, and super-resolution~\cite{SunB2021CVPR,LiF2023TMM,HuangY2022TPAMI,LeiX2026TIP}. Furthermore, degradation-aware prompts~\cite{Potlapalli2023NIPS,SunX2025TMM} are increasingly utilized to separate light scattering from the background scene. These prior-guided decoupling mechanisms provide crucial insights for flare removal tasks.
Chen~\etal~\cite{ChenG2025TCSVT} proposed the Location Prior Guidance (LPG) model, which estimates a continuous contamination map prior to the main network to focus processing on flare-affected regions. However, LPG guidance signals lack discriminability between flares and light sources, failing to instruct the main network to handle them distinctively, and are prone to misclassifying bright backgrounds as flares, introducing erroneous guidance.

\subsection{Vision Mamba} 

State Space Models (SSMs), notably Mamba~\cite{Albert2024COLM}, efficiently model long-range dependencies with linear complexity by unfolding 2D features into 1D sequences. Their data-dependent selective scan overcomes CNNs' spatial uniformity~\cite{JiangK2026TMM}, offering region-specific adaptability crucial for spatially varying degradations like scattering flares. However, applying SSMs to images necessitates effective scanning strategies to preserve structural integrity. Initial attempts, such as Vim~\cite{ZhuL2024ICML} and VMamba~\cite{LiuY2024NeurIPS}, utilized basic bi-directional scanning but struggled with fine-grained restoration. To address this, MambaIR~\cite{GuoH2024ECCV} integrates local convolutions with multi-directional SSMs to recover high-frequency details. Building upon this, DeflareMamba~\cite{HuangY2025ACMMM} leverages varied stride sampling to establish long-range pixel correlations within a hierarchical framework. Concurrently, EAMamba~\cite{Lin2025ICCV} introduces an all-around scanning multi-head mechanism. This efficiently aggregates diverse scanning sequences to capture holistic context, mitigating local pixel forgetting without adding computational overhead. 
Recently, MambaIRv2~\cite{GuoH2025CVPR} utilized semantic clustering to mitigate the causal limitations of standard SSMs. However, optical flares are uneven intensity degradations rather than distinct semantic objects, causing generic clustering to fail in capturing continuous variations. Building upon MambaIRv2, we instead exploit the physical causality of radial flare attenuation, leveraging specific priors to guide heterogeneous processing and significantly enhance restoration performance.
\section{METHOD}
This section details the proposed methodology, including the overall pipeline of DeflareMambav2, and its core modules.
\subsection{Overall Pipeline}

\begin{figure*}[!t]
\centering
\includegraphics[width=\linewidth]{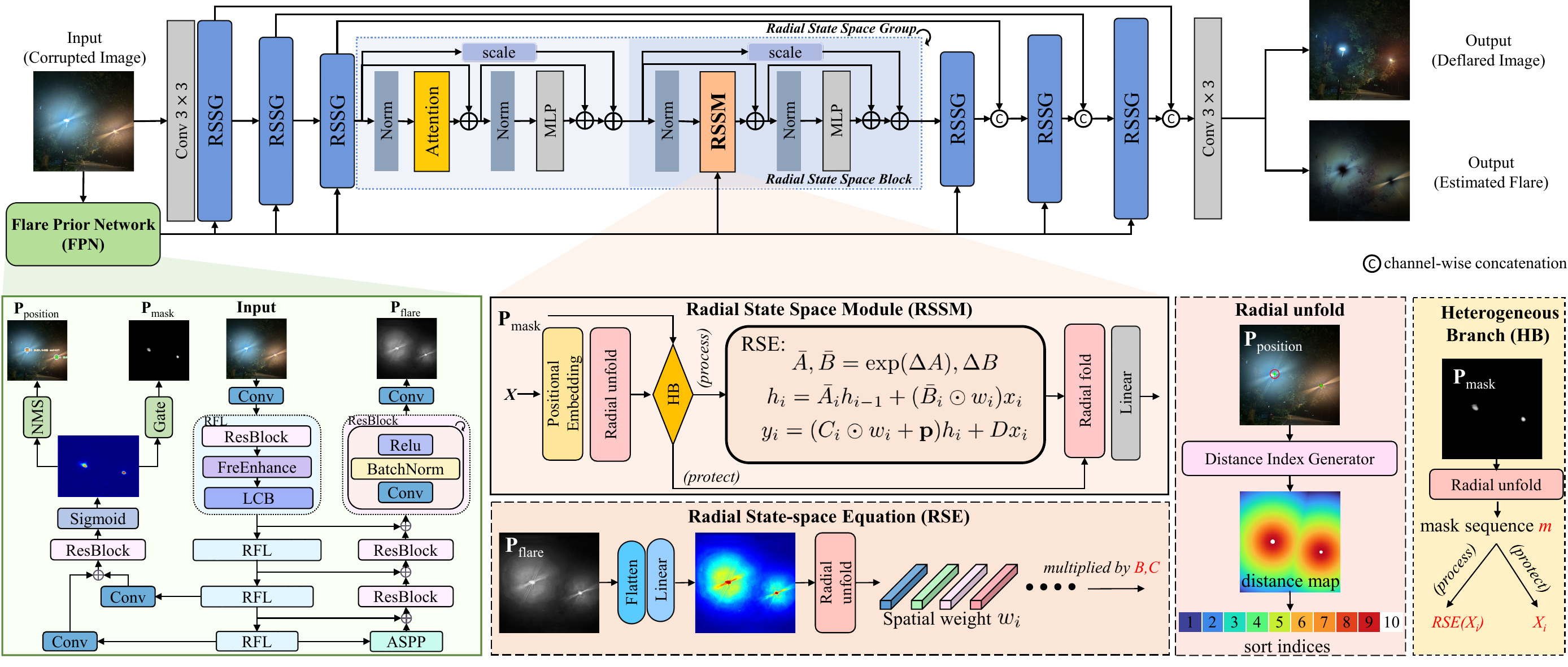}
\vspace{-1ex} 
\caption{The overall architecture of the proposed Mamba model for flare removal in night-time images (DeflareMambav2). 
The main network adopts a U-shaped architecture composed of seven RSSG modules.
The prior information extracted by the FPN is comprehensively integrated into the RSSM module through three distinct mechanisms.
(1) Radial unfold rearranges input features or priors based on sort indices. (2) Heterogeneous Branch (HB) acts as a spatial gating mechanism. (3) Radial State-Space Equation (RSE) leverages the acquired priors to dynamically modulate state parameters ($\bar{B}$ and $C$).}
\label{overpipline}
\end{figure*}

The overall pipeline of DeflareMambav2, illustrated in Fig.~\ref{overpipline}, consists of a main network and a Flare Prior Network (FPN). The main network is designed as a U-shaped hierarchical encoder-decoder architecture. Given a flare-corrupted input image $I_{\text{input}} \in \mathbb{R}^{H \times W \times 3}$, DeflareMambav2 first feeds $I_{\text{input}}$ into the FPN to obtain three priors. To provide corresponding spatial guidance, these priors are downsampled to match the feature map sizes, and then injected into the corresponding scales of the main network.

The main network comprises seven Radial State Space Groups (RSSGs), each consisting of two cascaded Radial State Space Blocks (RSSBs) which explicitly receive and utilize the prior information injected from the FPN. Specifically, the U-shaped architecture features a three-stage encoder, a symmetrical three-stage decoder, and a central bottleneck layer, all constructed upon RSSGs. 
Note that, an initial convolutional layer in encoder first extracts shallow features $I_0 \in \mathbb{R}^{H \times W \times C}$. Finally, given the ultimate feature map $I_7 \in \mathbb{R}^{H \times W \times C}$, a terminal convolutional layer projects it into a 6-channel output tensor $I_{\text{out}} \in \mathbb{R}^{H \times W \times 6}$. This tensor represents the concatenation of the flare-free image and the pure flare image.

\subsection{Flare Prior Network (FPN)}

Unlike random noise, scattering flares possess distinctive structural and physical priors. Exploiting these priors is essential to guide the restoration network, enabling it to aggressively remove artifacts while preserving background fidelity. 
To achieve this, our FPN explicitly models the flare degradation by estimating three critical guidance signals. First, the FPN precisely localizes the light source coordinates ($P_{position}$). This exact localization establishes the foundational prerequisite for the Radial Decay physical prior utilized in the subsequent SSM module. Second, since standard feature extraction is spatially invariant, the FPN generates a discrete source mask ($P_{mask}$) to break this uniformity. This mask allows the network to apply distinct strategies: preserving the light source and removing the flares. Finally, to address the highly spatially varying nature of flare corruption, the FPN estimates a continuous contamination map ($P_{flare}$). This continuous guidance empowers the main network to adaptively modulate its restoration intensity at the pixel level.

The overall architecture of the proposed FPN, illustrated in Fig.~\ref{overpipline}, adopts a hierarchical encoder structure built upon RFL components, which progressively downsamples the spatial resolution to extract multi-scale features. These representations are subsequently routed into two task-specific branches. The Flare Head reconstructs the full-resolution, pixel-wise contamination map $P_{flare}$. Simultaneously, the Light Head predicts the source probability map. This probability map is then processed via confidence thresholding to yield the binary segmentation mask $P_{mask}$, and further refined using Non-Maximum Suppression (NMS) to pinpoint the precise source coordinates $P_{position}$. Consequently, the Prior Network serves as a lightweight guide focused on macroscopic geometry and illumination rather than fine details. By employing a multitask architecture, it predicts a unified contamination map for both flares and light sources, alongside a specific source mask to isolate them. This dual prediction strategy bypasses the inherent difficulty of decoupling entangled flares and sources, providing precise modeling guidance to the main network.

While standard convolutional networks adequately localize light sources and estimate generic flare distributions, they struggle to isolate streak flares, which cause severe degradation due to their extensive anisotropic and multiscale geometries. To overcome this limitation, we introduce two specialized components within the RFL module, including Frequency Enhancement (FreEnhance) module and Line Capture Block (LCB). Specifically, since streak flares predominantly manifest as high-frequency signals \cite{DengH2024TIP}, the FreEnhance module introduces learnable parameters to adaptively amplify these high-frequency components in the frequency domain. Concurrently, the LCB employs specialized horizontal and vertical convolutions, synergized with deformable convolutions \cite{DaiJ2017ICCV}, to comprehensively aggregate anisotropic features across varying orientations. Integrating these modules enables the FPN to enhance long-range representations, achieving precise capture and robust prediction of complex streak flares.

Ultimately, while maintaining a highly lightweight footprint of merely 1.65M parameters, our FPN accurately deduces $P_{position}$, $P_{mask}$, and $P_{flare}$ as the essential physical guidance for flare removal. To fully harness these extracted priors, we design three synergistic modules within the main network.

\subsection{Radial State Space Module (RSSM)}

As the central contribution of this work, the Radial State Space Module (RSSM) forms the core of our backbone, integrating three tailored components within the SSM framework to explicitly exploit the three priors generated by the FPN. To maintain spatial consistency, the priors are downsampled and injected into the RSSM as aligned guidance. Structurally, the RSSM comprises a front-end Attention module~\cite{ZeL2021ICCV} and a prior-enhanced SSM. Unlike SGN-unfold~\cite{GuoH2025CVPR}, which relies on weak spatial causality ill-suited to the minimal semantic response of flares, the SSM leverages the injected priors for targeted restoration of strongly causal flare sequences through three sequential operations. Radial unfold first exploits $P_{position}$ to generate spatial sorting indices for feature map unfolding, establishing a global traversal from weakly to heavily contaminated regions. The Heterogeneous Branch (HB) then employs $P_{mask}$ to explicitly identify source pixels and bypass SSM processing. Finally, the Radial State-space Equation (RSE) utilizes $P_{flare}$ to dynamically modulate restoration intensity at pixel-wise granularity, after which the processed sequence is folded back to restore the original spatial layout.

\subsubsection{Radial unfold}
\begin{figure}[t]
    \centering

        
        \includegraphics[width=0.5\linewidth]{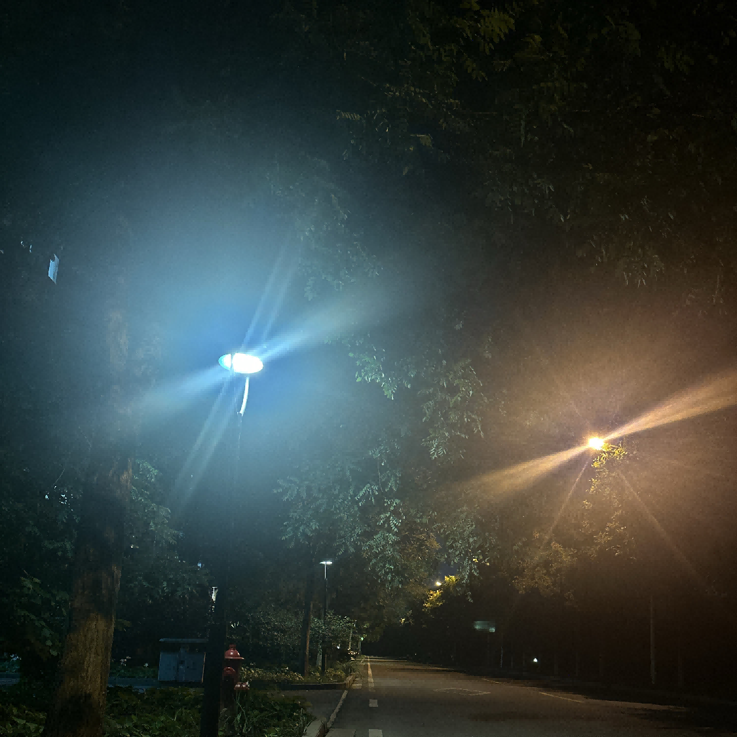}\hfill
        \includegraphics[width=0.5\linewidth]{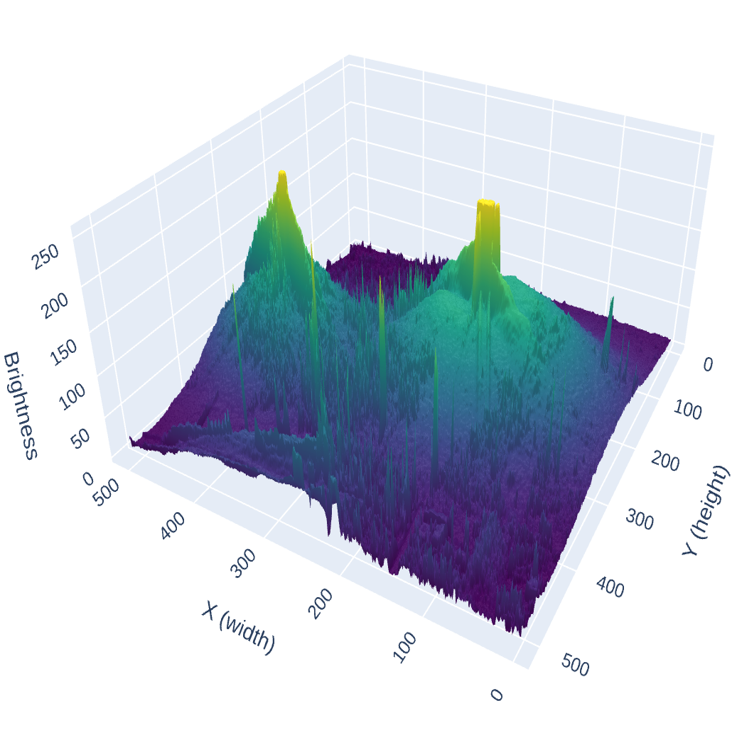}
    \vspace{-0.2cm}
    \caption{Visualization of the radial attenuation characteristics of flares. 
    This radial decay pattern is consistently observed in authentic captured scenes. Excluding the fluctuating intensities of streak flares, the primary contamination strictly adheres to radial attenuation. Furthermore, since streak flares inherently decay radially along their specific trajectories, the overall composite flare universally conforms to this spatial attenuation property.}
    \label{fig:radial_attenuation}
\end{figure}
Governed by the physics of optical scattering, flare-induced degradation is highly non-uniform, characterized by pronounced radial attenuation from the light source centers, as illustrated in Fig.~\ref{fig:radial_attenuation}. While distant regions remain largely intact with rich high-frequency details, areas proximate to the sources suffer extreme contamination, pixel saturation, and severe information loss. Despite this strong physical prior, existing CNNs and ViTs typically process degraded images as isotropic 2D grids using uniform spatial windows. This spatially agnostic paradigm forces networks to compromise across varying degradation levels, which inevitably blurs clean backgrounds and leaves residual artifacts in heavily corrupted regions.

To overcome this limitation, we introduce the Radial unfold strategy. It leverages sort indices to synchronously flatten the input feature map ${X}$, ${P}_{\text{mask}}$ and ${P}_{\text{flare}}$ into aligned sequences under a unified spatial transformation. To derive these indices, we explicitly model the spatial degradation prior of lens flares via a global distance map $\mathbf{D} \in \mathbb{R}^{H \times W}$. Assuming a set of $K$ detected light source centers $\mathcal{S} = \{\mathbf{c}_k\}_{k=1}^K$, the corruption intensity at any spatial location $p=(x, y)$ is generally governed by its closest source. Thus, we formulate $\mathbf{D}$ via a non-linear minimum aggregation of Euclidean distances:

\begin{equation}
    \mathbf{D}({p}) = \min_{\mathbf{c}_k \in \mathcal{S}} \|{p} - \mathbf{c}_k\|_2
    \label{eq:distance_map}
\end{equation}

For the scenes where the set of light sources $\mathcal{S}$ is empty, we define the distance $\mathbf{D}({p})$ as approaching infinity. This naturally ensures operational robustness.
Instead of relying on conventional raster scanning, we derive a spatial permutation function $\pi$ over the 2D spatial grid. This function maps the sequential 1D indices to 2D spatial coordinates by sorting them descendingly according to $\mathbf{D}$:

\begin{equation}
    \mathbf{D}(\pi(1)) \ge \mathbf{D}(\pi(2)) \ge \dots \ge \mathbf{D}(\pi(L))
    \label{eq:sorting}
\end{equation}

Given an input feature tensor ${X} \in \mathbb{R}^{H \times W \times d}$, we formulate the recombined sequence ${x} \in \mathbb{R}^{L \times d}$ via the mapping ${x}[i] = {X}(\pi(i))$. This degradation-aware transformation systematically reorganizes spatial features from the furthest flare-free areas to the nearest heavily corrupted regions. As a result, it establishes an explicit and physically meaningful trajectory for subsequent sequence modeling.

\subsubsection{Heterogeneous Branch (HB)}

In global sequence modeling, scanning trajectories inevitably traverse light source regions. Since light source pixels entail fundamentally different restoration objectives from flare-corrupted areas, homogeneous blind processing forces the State Space Model (SSM) to accommodate conflicting objectives simultaneously. The saturated intensities of light sources thus induce a severe memory washout effect, corrupting the contextual priors accumulated in the hidden state ${h}_t$ and disrupting coherent flare artifact elimination. To address this, we propose a source-aware protective routing mechanism. The predicted binary mask $P_{\mathrm{mask}}$ is flattened via Radial unfold into a sequence $\{m\} \in \{0, 1\}^{L}$, ensuring precise alignment with the input feature sequence $x \in \mathbb{R}^{L \times d}$. The routing for the $i$-th token is formulated as:
\begin{equation}
{y_i} = \begin{cases}
x_i, & {\text{if}}\ {m_i = 1} \\
\textit{RSE}(x_i), & {\text{otherwise.}}\
\end{cases}
\label{eq:routing}
\end{equation}

Unlike conventional soft-masking operations, we decouple the sequence via index-based feature extraction. Clean background tokens and flare-corrupted tokens are fed into the RSE for memory accumulation or restoration, while the protected light source tokens are preserved and bypass the RSE. Upon completion, the processed tokens and the skipped source tokens are inversely mapped back to their original spatial coordinates. This explicit routing strategy inherently decouples the optimization objective. Consequently, it compels SSM to dedicate its entire representational capacity exclusively to the structural restoration of flare-corrupted regions.


\subsubsection{Radial State-space Equation (RSE)}

While \textit{Radial unfold} establishes a global weak-to-severe scanning trajectory, flare degradation inherently exhibits severe spatial heterogeneity at the local pixel level. Specifically, streak flares possess drastically higher intensities and abrupt structural transitions compared to the surrounding diffuse veiling glare. To accommodate such drastic local variations, we introduce the Radial State-Space Equation (RSE) to complement the macroscopic routing with pixel-level fine-grained feature modulation, adaptively regulating the state-space memory dynamics.

Initially, following Mamba~\cite{Albert2024COLM}, we discretize the SSM via the ZOH rule, yielding the discrete state transition and output:

\begin{equation}  
\begin{aligned}  
    {h}_i &= {\bar{A}}_i {h}_{i-1} + {\bar{B}}_i {x}_i \\
    {y}_i &= {C}_i {h}_i + {D} {x}_i
\end{aligned}  
\label{eq:discrete_ssm}  
\end{equation}  

Following the Flare7K++ training protocol, we formulate flare removal as an explicit additive decomposition task, separating inputs into clean and flare components. Guided by this physical concept, we propose an energy-driven synchronous excitation mechanism. Specifically, the FPN estimates a spatial contamination prior $P_{\text{flare}}$ where higher values indicate severe degradation. This prior generates a weight $W$ to amplify the state space parameters:

\begin{equation}
\begin{aligned}
    W &= \Phi\Big(\text{Linear}\big(\text{Flatten}\big(P_{\mathrm{flare}}\big)\big)\Big), \\
\end{aligned}
\label{eq:spatial_weights}
\end{equation}
where $\text{Linear}(\cdot)$ denotes a linear layer projecting the 1D input into a $d$-dimensional hidden space, and $\Phi(\cdot)$ represents the Sigmoid function. To preserve spatial alignment, $W$ is similarly flattened into $w$ via Radial unfold. These generated excitations directly modulate the core state-space parameters. The overall forward pass of our RSE is thus formulated as:

\begin{equation}
\begin{gathered}
\bar{A},\bar{B} = \exp(\Delta A), \Delta B \\
h_i = \bar{A}_i h_{i-1} + (\bar{B}_i \odot w_i) x_i \\
y_i = (C_i \odot w_i + \mathbf p) h_i + D x_i
\end{gathered}
\end{equation}
where $\odot$ denotes the Hadamard product, and $p$ represents an optional additive token prompt for further feature refinement. Also projected from $P_{flare}$, $\mathbf p$ is designed to act synchronously with $w_i$ at the same time step $i$.

Unlike traditional noise suppression, our task requires explicitly subtracting flare energy from the blended observation. The proposed synchronous amplification aligns perfectly with this additive decomposition paradigm. In heavily corrupted regions where the spatial prior ${P}_{\text{flare}}$ approaches $1$, extreme optical aberrations dominate the visual signal. Amplifying both $\bar{B}_i$ and $C_i$ forces the network to concentrate its representational bandwidth on these high-energy areas. Specifically, an enlarged $\bar{B}_i$ effectively captures severe flare structures, while an amplified $C_i$ ensures their precise reconstruction.

Working synergistically with the HB, which explicitly breaks spatial uniform processing, the network effectively isolates extreme light source features. Consequently, even when ${P}_{\text{flare}}$ approaches its maximum value to yield peak excitation at the light source, the HB prevents these extreme features from polluting the state-space memory in the RSE. 

\section{Experiment}

\subsection{Experimental Setup}

\subsubsection{Datasets} 
We train the main network on the Flare7K++ dataset~\cite{DaiY2024TPAMI}, dynamically constructing training pairs by sampling backgrounds from the 24K Flickr dataset and stochastically retrieving flare-source pairs from Flare7K~\cite{DaiY2022NeurIPS} and Flare-R~\cite{DaiY2024TPAMI}. Standard augmentations following~\cite{DaiY2024TPAMI, HuangY2025ACMMM, JieZ2025ICCV, MaT2025TASE} are applied to the training samples to simulate unconstrained capturing conditions.

However, the standard pipeline limits each training sample to a single light source, deviating from real nighttime scenes with multiple distributed sources. Multiple intense light beams induce non-linear phenomena such as sensor saturation and inter-lens scattering~\cite{zhouY2023ICCV, DaiY2024TPAMI}, which simple additive synthesis cannot faithfully replicate. Such imperfect multi-source data is insufficient to supervise the main network, yet adequate to train the lightweight FPN to deliver robust prior guidance in complex multi-source scenes.

Specifically, given a clean background and sampled flare and light images, we apply independent random affine transformations to each pair. The transformed components are then accumulated via blending to generate a composite flare. Subsequently, the degraded input is synthesized by blending the background with composite flare. Finally, the composite flare image is used as the ground truth (GT) for the flare prediction part in the FPN. Furthermore, to provide precise localization cues, we formulate light source detection as a keypoint estimation task ~\cite{ZhouX2019CVPR, HeiL2020IJCV} and generate a GT spatial heatmap $H$ via a 2D Gaussian kernel. Consequently, $H_p \in [0, 1]$ represents the exact confidence score of a light source being present at pixel $p$, serving as the pixel-wise supervision signal for the subsequent heatmap loss.

\begin{figure}[!t]
    \centering

    \includegraphics[width=\linewidth]{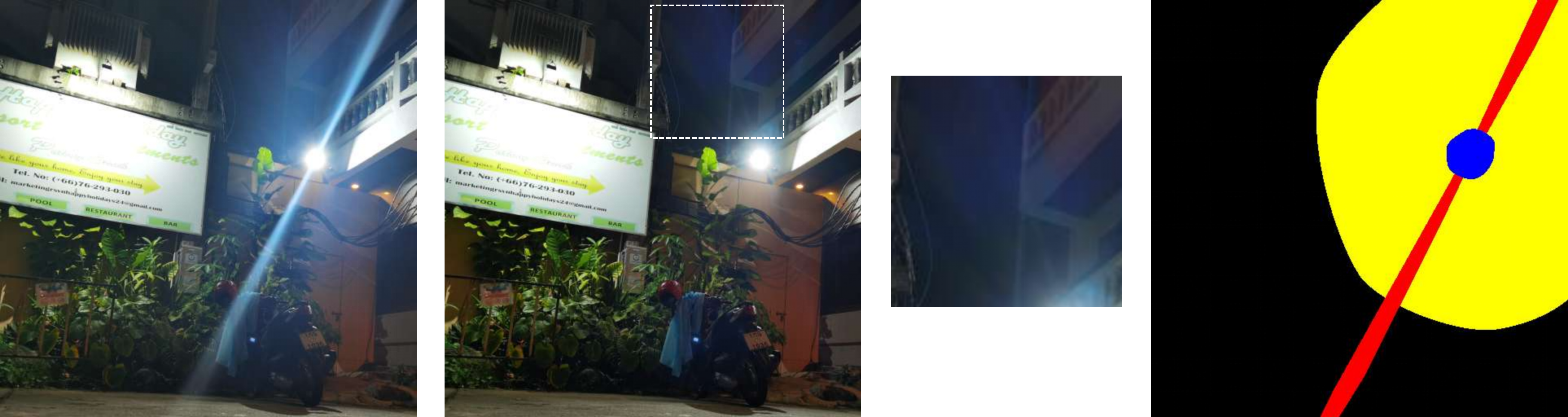}
    \\ \vspace{0.5ex}
    {\small (a) Residual flares in Flare7K-Real GT} 
    \\ \vspace{1.5ex} 
    
    \includegraphics[width=\linewidth]{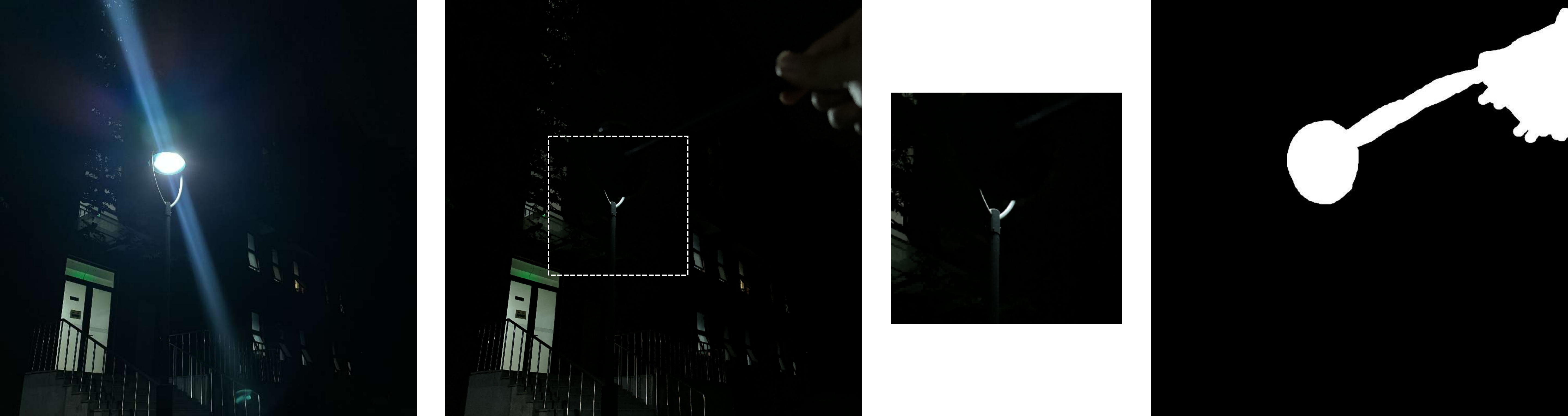}
    \\ \vspace{0.5ex}
    {\small (b) Missing light sources in FlareX GT} 
    
    \caption{Visualizations of evaluation benchmarks. From left to right: input, GT, flaw region, and evaluation mask.}
    \label{fig:benchmarks_vis}
\end{figure}

\subsubsection{Evaluation Metrics and Benchmarks}

To enable rigorous assessment, we conduct experiments on two test sets, Flare7K-real~\cite{DaiY2022NeurIPS} and FlareX~\cite{QuL2025NeurIPS}, each comprising 100 real-world image pairs, and introduce two region-specific metrics. Light-PSNR is computed within the light source mask of Flare7K-real (\ie, the blue region in Fig.~\ref{fig:benchmarks_vis}(a)) to quantify source structure fidelity. Clean-PSNR is computed within the uncovered mask of FlareX (the black region in Fig.~\ref{fig:benchmarks_vis}(b)) to evaluate flare removal capability while bypassing residual flare biases in Flare7K-real.

Both benchmarks have inherent limitations. Flare7K-real's GT is obtained via mechanical lens wiping, potentially leaving residual artifacts, while FlareX's physical occlusion guarantees flare-free backgrounds but inevitably corrupts the source region in its GT. Consequently, Clean-PSNR/SSIM on FlareX provides the most reliable measure of flare removal, while Light-PSNR on Flare7K-real reflects light source preservation. We also report standard metrics including PSNR, SSIM~\cite{WangZ2004TIP}, LPIPS~\cite{zhang2018CVPR}, S-PSNR, and G-PSNR~\cite{DaiY2024TPAMI} on Flare7K-real as complementary references. These global metrics are excluded from FlareX evaluation, as partial background loss in its GT renders full-reference metrics unreliable.

\subsubsection{Implementation Details}
The proposed DeflareMambav2 is optimized using the Adam~\cite{KingmaD2015ICLR} optimizer with a batch size of $1$. To synthesize training samples of size 512×512, we utilize the Flickr24K dataset as pristine backgrounds. These backgrounds are subsequently blended with real and synthetic flare components from the Flare7K++ dataset at an equal ratio of $1:1$. The network is trained for $600,000$ iterations. The initial learning rate is set to $1 \times 10^{-4}$, which is halved at $200,000$ iterations and maintained constant for the remainder of the training process. All experiments are conducted on a single NVIDIA GeForce RTX 5090D GPU.

\subsection{Loss Function}
To ensure stable convergence and prevent complex restoration gradients from corrupting the learned spatial priors, we adopt a decoupled two-stage training strategy.
\subsubsection{Loss of FPN} 

In this stage, the FPN exclusively predicts the continuous flare intensity 
and the spatial light source heatmap. The total loss $\mathcal{L}_{FPN}$ is defined as:
\begin{equation}
    \mathcal{L}_{FPN} = \underbrace{\mathcal{L}_{char} + \mathcal{L}_{err} + 
    \mathcal{L}_{weak}}_{\mathit{flare}} + \underbrace{\mathcal{L}_{hm}}_{\mathit{source}}
\end{equation}

For global flare intensity prediction, standard Binary Cross-Entropy (BCE) heavily penalizes intermediate values, making it suboptimal for continuous intensity regression. To restore global degradation structure while preserving highlight boundaries, we hierarchically construct $\mathcal{L}_{flare}$ from three complementary losses. $\mathcal{L}_{char}$ applies a Charbonnier penalty~\cite{laiW2017CVPR} between the predicted flare and ground truth to ensure global structural fidelity. $\mathcal{L}_{err}$ employs an adaptively weighted BCE loss modulated by a bounded error map $W_{err}$ with a stop-gradient operation to focus on hard examples while stabilizing optimization. $\mathcal{L}_{weak}$ imposes an L1 penalty restricted to low-intensity regions, preventing over-suppression of dim scattered flares caused by vanishing gradients.

Furthermore, to mitigate the severe spatial class imbalance inherent in light source localization, we apply a penalty-reduced focal loss~\cite{HeiL2020IJCV} for heatmap estimation:
\begin{equation}
    \mathcal{L}_{hm} = \frac{-1}{N} \Bigg[ \sum_{p \in \mathcal{P}^+} (1 - \hat{H}_p)^\alpha \log(\hat{H}_p) + \sum_{p \in \mathcal{P}^-} w_p \hat{H}_p^\alpha \log(1 - \hat{H}_p) \Bigg]
\end{equation}

where $N$ is the number of valid light sources, and $\mathcal{P}^+$ and $\mathcal{P}^-$ denote the positive and negative pixel sets, respectively. We set the focusing parameter $\alpha=2$. The distance-aware weight $w_p = (1 - H_p)^\beta$ (with $\beta=4$) relaxes the penalty for near-center false positives, smoothly guiding the prediction toward true centers.

\subsubsection{Loss of the main network} Once the FPN converges, its parameters are strictly frozen. The primary network is then optimized to recover the flare-free image by leveraging the explicit spatial priors generated by the FPN.

Following previous methods \cite{DaiY2024TPAMI, Kotp2024ICASSP, HuangY2025ACMMM}, the main network is supervised by a standard combination of reconstruction and perceptual metrics. Specifically, the predicted deflare image and flare image are explicitly constrained against their ground-truth counterparts via $\mathcal{L}_1$ and VGG-based perceptual loss $\mathcal{L}_{vgg}$ ~\cite{johnson2016ECCV}. Furthermore, $\mathcal{L}_{rec}$ is introduced to enforce physical consistency between the recomposed predictions and the degraded input image. The total objective is formulated as:
\begin{equation}
    \mathcal{L}_{main} = \mathcal{L}_{1} + \mathcal{L}_{vgg} + \mathcal{L}_{rec}
\end{equation}
where all balancing weights are empirically set to equal, which is strictly consistent with the established configurations in Flare7K++~\cite{DaiY2024TPAMI}.

\subsection{Comparison With Previous Methods}
\begin{table*}[t]
    \centering
    \caption{Quantitative Comparison of Flare Restoration Performance on the FlareX and Flare7K-Real Datasets, with the Best and Second-Best Scores Highlighted in \textbf{Bold} and \underline{Underlined}, Respectively}
    \label{tab:sota_comparison}

    \resizebox{\textwidth}{!}{
    \begin{tabular}{l cc cccccc c} 
        \toprule
        \multirow{2}{*}{Method} & \multicolumn{2}{c}{FlareX \cite{QuL2025NeurIPS}} & \multicolumn{6}{c}{Flare7K-real \cite{DaiY2022NeurIPS}} & \multirow{2}{*}{Params (M) $\downarrow$} \\
        \cmidrule(lr){2-3} \cmidrule(lr){4-9} 
        & Clean-PSNR $\uparrow$ & Clean-SSIM $\uparrow$ & Light-PSNR $\uparrow$ & PSNR $\uparrow$ & SSIM $\uparrow$ & LPIPS $\downarrow$ & G-PSNR $\uparrow$ & S-PSNR $\uparrow$ & \\
        \midrule
        Wu et al. \cite{WuY2021ICCV}        & -- & -- & -- & 24.613 & 0.871 & 0.0598 & 21.77 & 16.73 & 34.53 \\
        Flare7K \cite{DaiY2022NeurIPS}      & 28.250 & 0.760 & 26.54 & 24.925 & 0.890 & 0.0554 & 22.51 & 21.12 & 20.55 \\
        Kotp et al. \cite{Kotp2024ICASSP}   & 28.202 & 0.760 & 27.86 & 27.662 & 0.898 & \underline{0.0422} & 23.99 & 22.85 & 129.21 \\
        Flare7K++ \cite{DaiY2024TPAMI}      & 28.071 & 0.761 & 27.80 & 27.633 & 0.895 & 0.0428 & 23.95 & 22.60 & 20.55 \\
        DeflareMamba \cite{HuangY2025ACMMM} & \underline{29.662} & 0.774 & 27.96 & 27.787 & 0.898 & 0.0450 & 24.41 & 22.38 & \underline{14.11} \\
        FGRNet \cite{JieZ2025ICCV}          & 28.784 & 0.758 & 27.98 & \underline{28.053} & 0.898 & 0.0424 & \underline{24.49} & \underline{23.16} & 25.99 \\
        SGSFT \cite{MaT2025TASE}            & 29.247 & \underline{0.776} & \textbf{28.25} & \textbf{28.078} & \textbf{0.904} & \textbf{0.0416} & 24.48 & \textbf{23.31} & 54.45 \\
        \midrule 
        \textbf{Ours}                       & \textbf{30.099} & \textbf{0.779} & \underline{28.04} & 27.861 & \underline{0.900} & 0.0442 & \textbf{24.66} & 22.85 & \textbf{8.31+1.65} \\
        \bottomrule
    \end{tabular}
    } 
\end{table*}
To validate the superiority of DeflareMambav2, we conduct comprehensive evaluations against leading SOTA flare removal approaches. The compared baselines span from early representative works, including Wu~\cite{WuY2021ICCV} and the unified Flare7K~\cite{DaiY2022NeurIPS}, to recent accessible frameworks such as the Flare7K++ baseline~\cite{DaiY2024TPAMI} and the dual phase method by Kotp and Torki~\cite{Kotp2024ICASSP}. We additionally evaluate the recent DeflareMamba~\cite{HuangY2025ACMMM}, the Self Prior Guided Spatial and Fourier Transformer (SGSFT)~\cite{MaT2025TASE}, alongside Flare and Glare Removal Network (FGRNet)~\cite{JieZ2025ICCV}. Detailed quantitative results are summarized in Table~\ref{tab:sota_comparison}.

\begin{figure*}[htbp]
    \centering
    \setlength{\tabcolsep}{1pt} 
    \begin{tabular}{cccccccc}
        
        \includegraphics[width=0.12\linewidth]{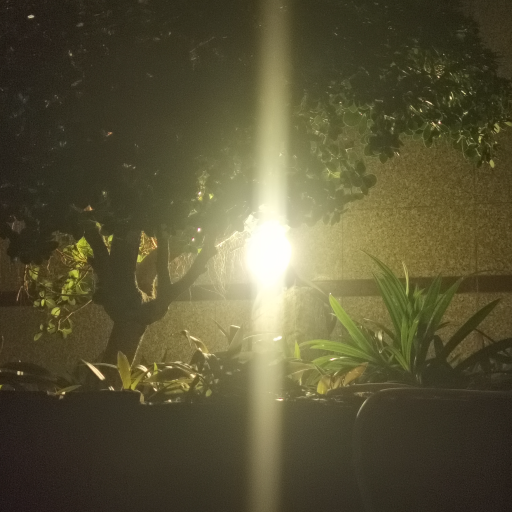} &
        \includegraphics[width=0.12\linewidth]{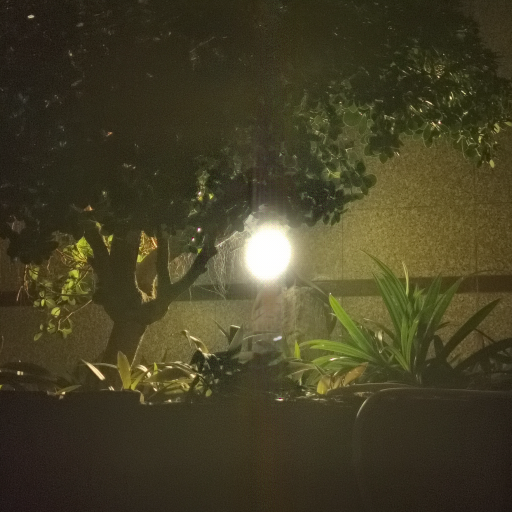} &
        \includegraphics[width=0.12\linewidth]{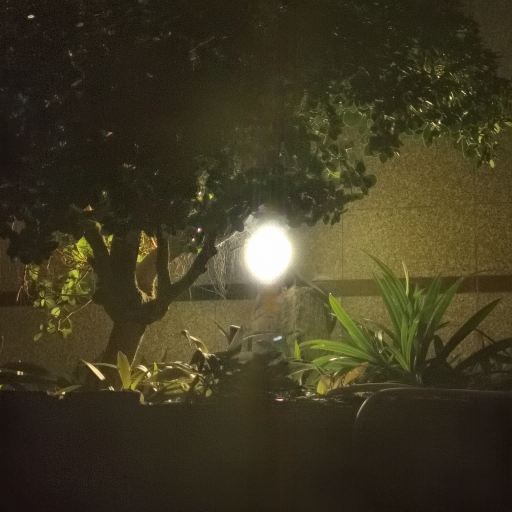} &
        \includegraphics[width=0.12\linewidth]{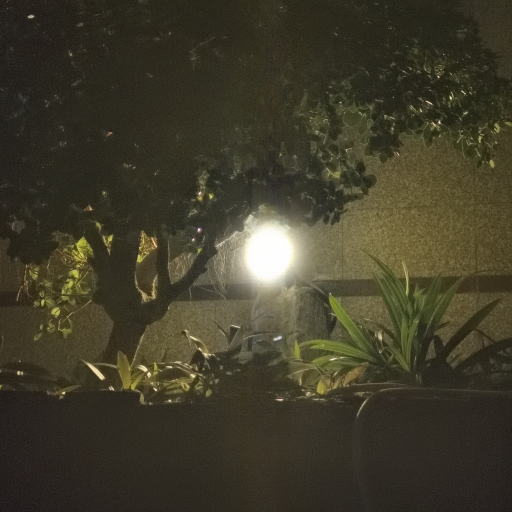} &
        \includegraphics[width=0.12\linewidth]{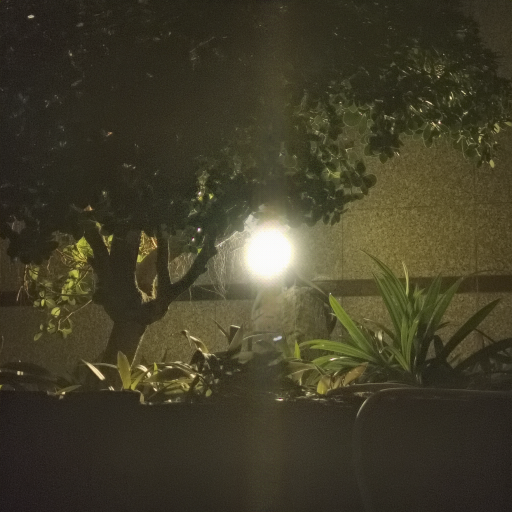} &
        \includegraphics[width=0.12\linewidth]{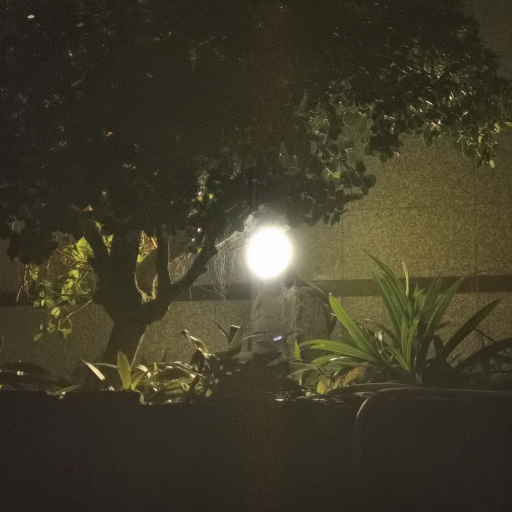} &
        \includegraphics[width=0.12\linewidth]{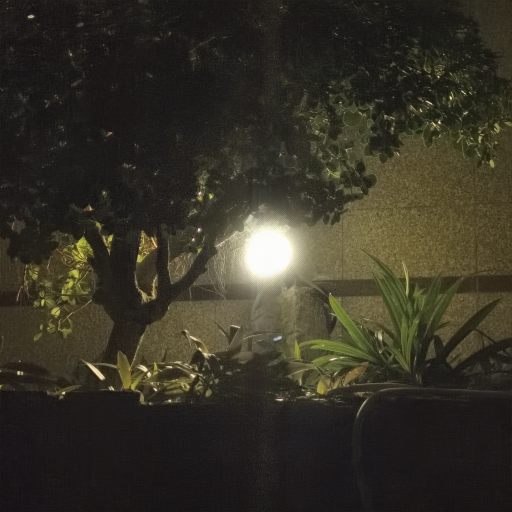} &
        \includegraphics[width=0.12\linewidth]{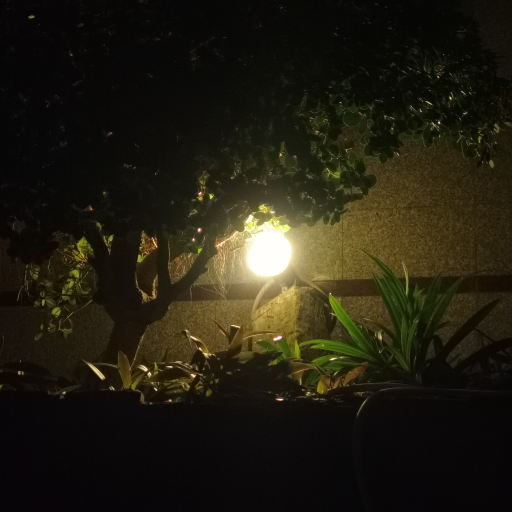} \\
        
        \includegraphics[width=0.12\linewidth]{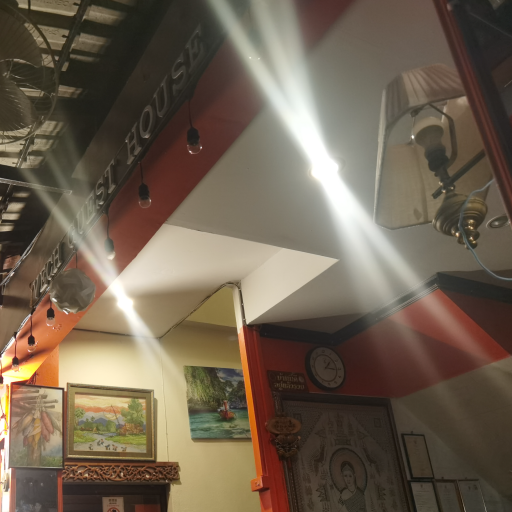} &
        \includegraphics[width=0.12\linewidth]{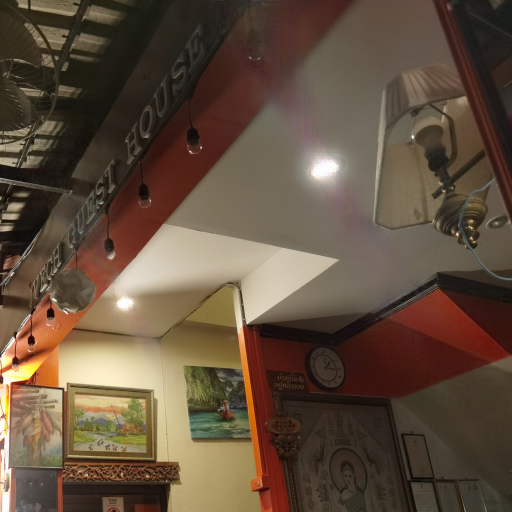} &
        \includegraphics[width=0.12\linewidth]{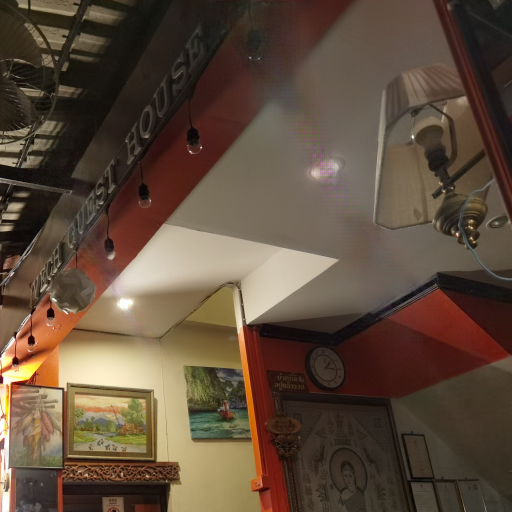} &
        \includegraphics[width=0.12\linewidth]{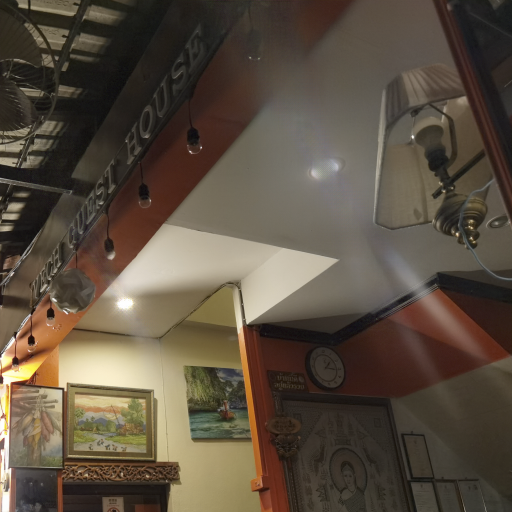} &
        \includegraphics[width=0.12\linewidth]{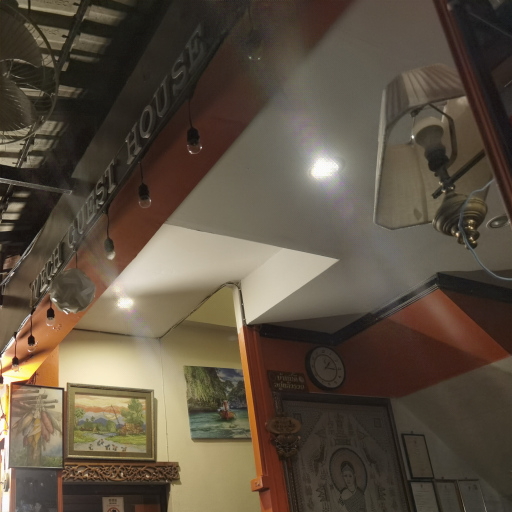} &
        \includegraphics[width=0.12\linewidth]{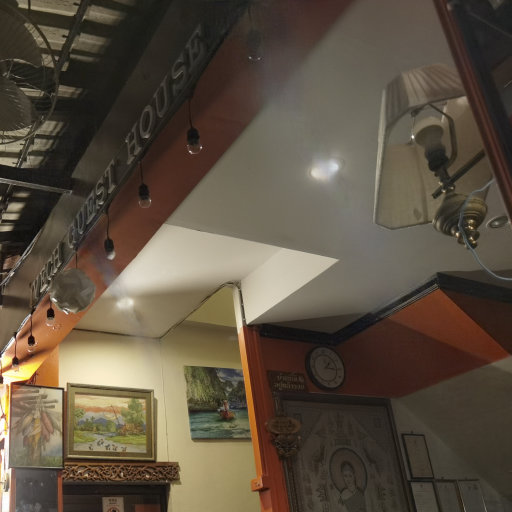} &
        \includegraphics[width=0.12\linewidth]{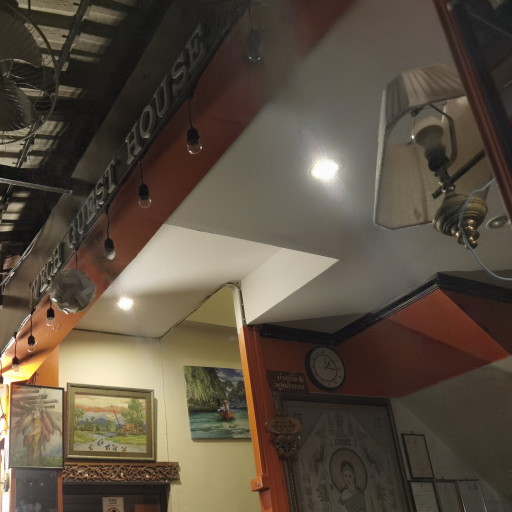} &
        \includegraphics[width=0.12\linewidth]{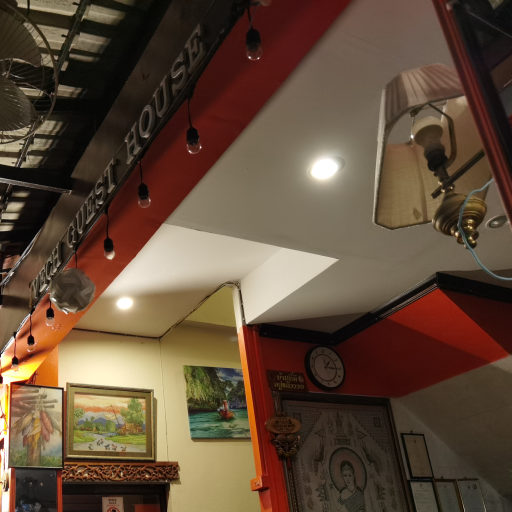} \\
        
        \includegraphics[width=0.12\linewidth]{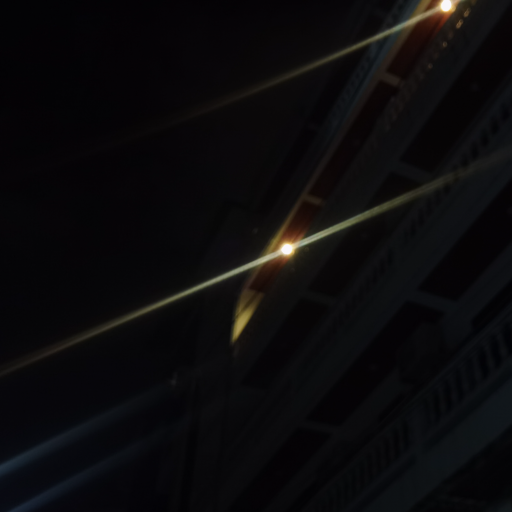} &
        \includegraphics[width=0.12\linewidth]{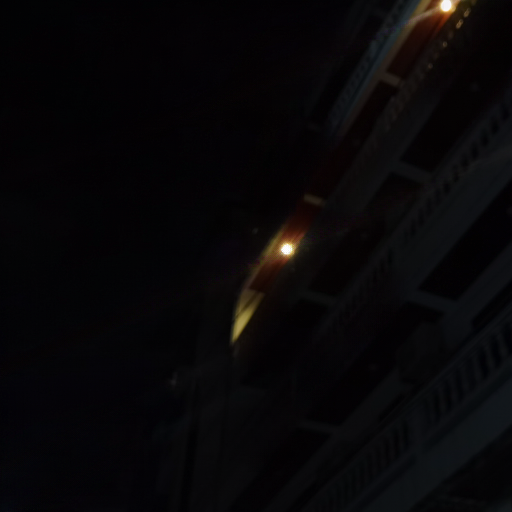} &
        \includegraphics[width=0.12\linewidth]{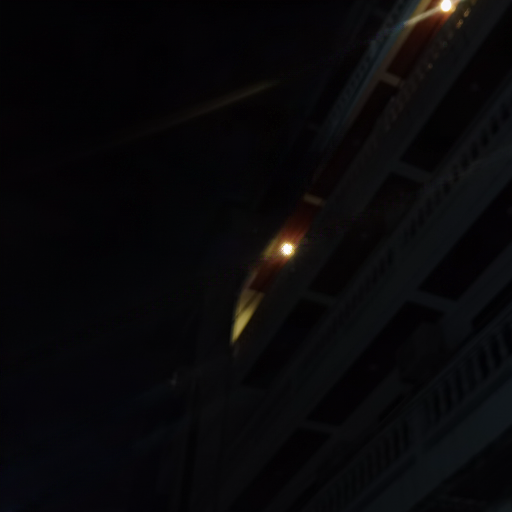} &
        \includegraphics[width=0.12\linewidth]{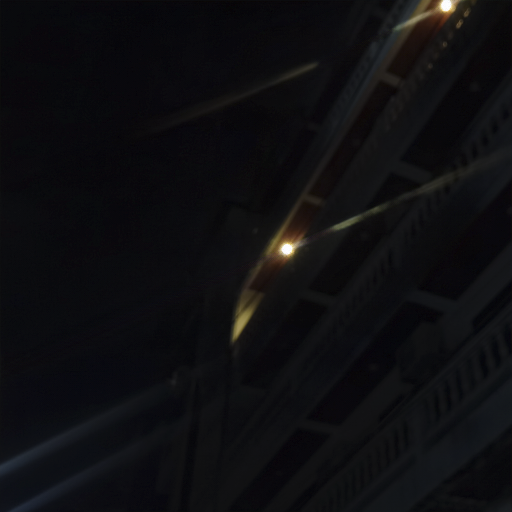} &
        \includegraphics[width=0.12\linewidth]{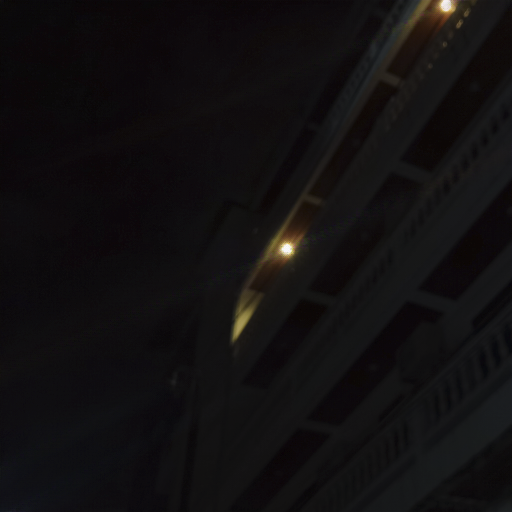} &
        \includegraphics[width=0.12\linewidth]{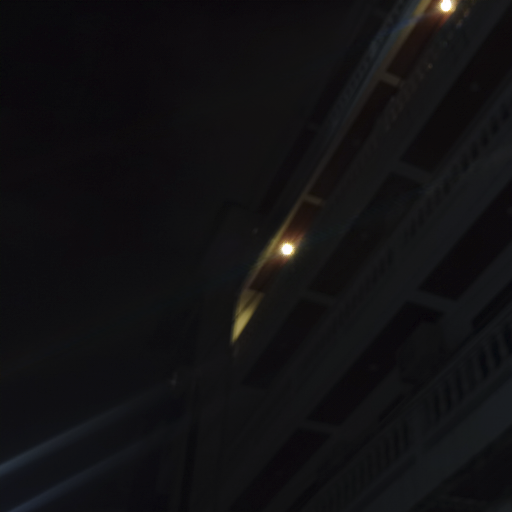} &
        \includegraphics[width=0.12\linewidth]{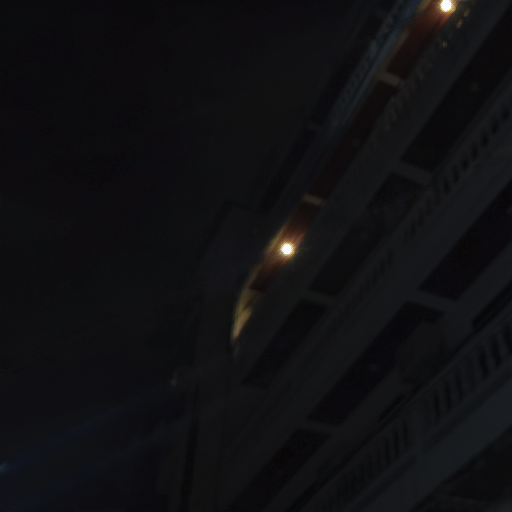} &
        \includegraphics[width=0.12\linewidth]{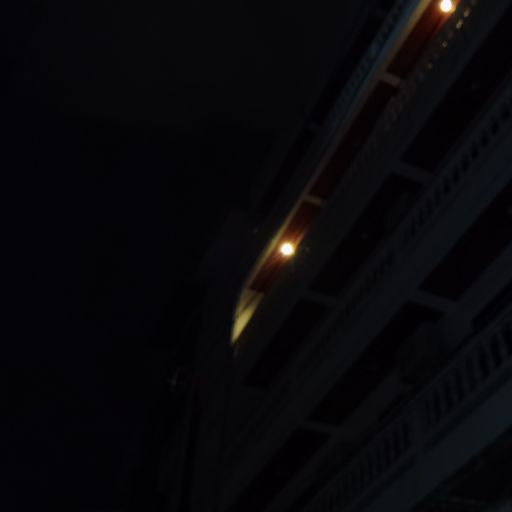} \\
        
        \includegraphics[width=0.12\linewidth]{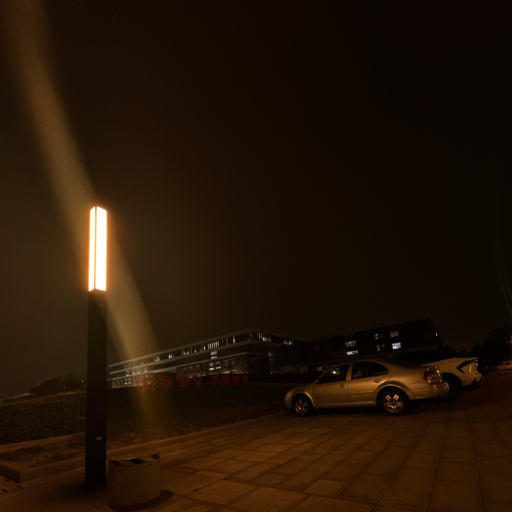} &
        \includegraphics[width=0.12\linewidth]{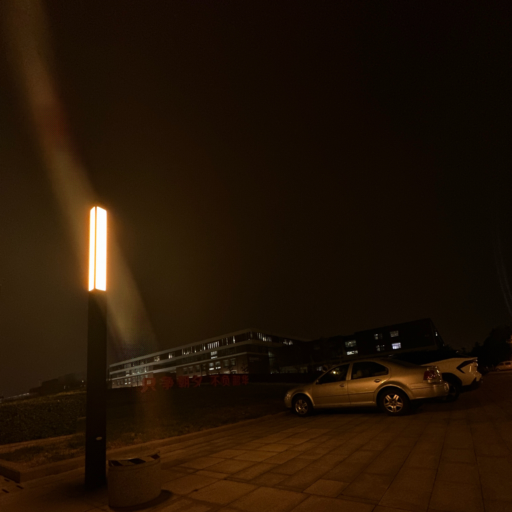} &
        \includegraphics[width=0.12\linewidth]{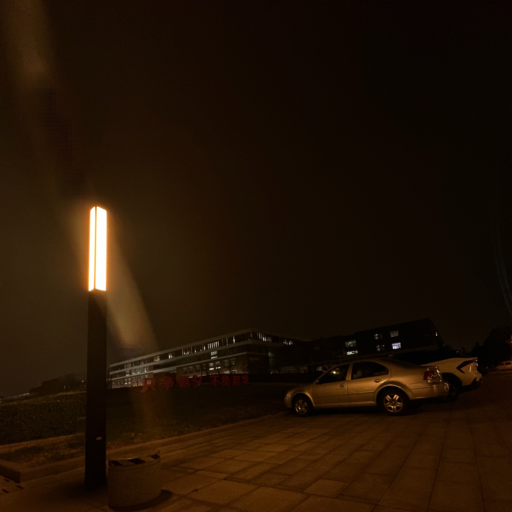} &
        \includegraphics[width=0.12\linewidth]{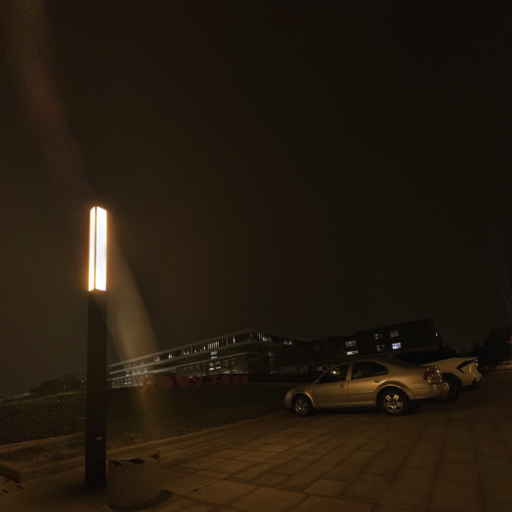} &
        \includegraphics[width=0.12\linewidth]{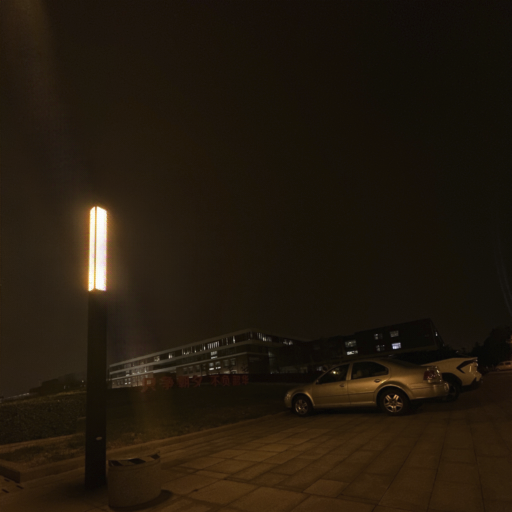} &
        \includegraphics[width=0.12\linewidth]{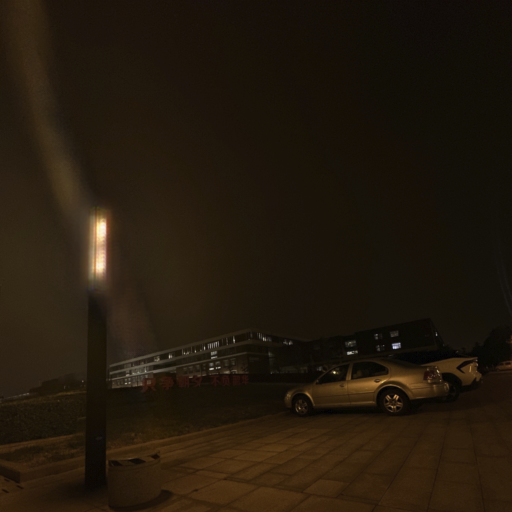} &
        \includegraphics[width=0.12\linewidth]{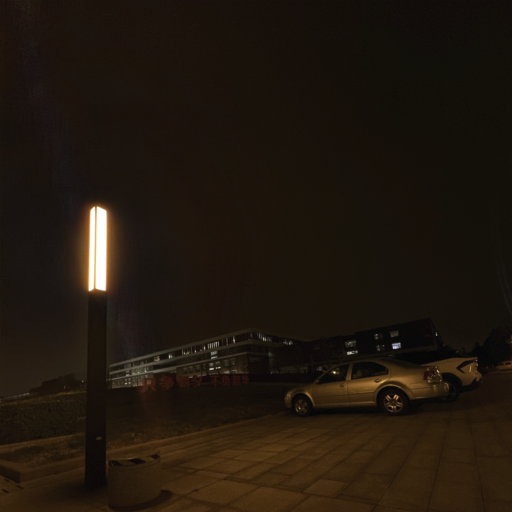} &
        \includegraphics[width=0.12\linewidth]{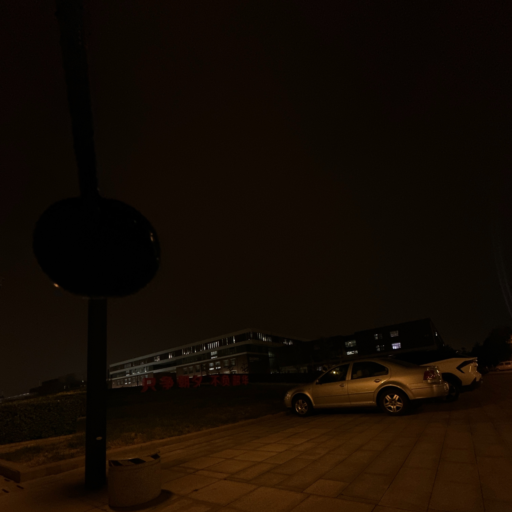} \\
        
        \includegraphics[width=0.12\linewidth]{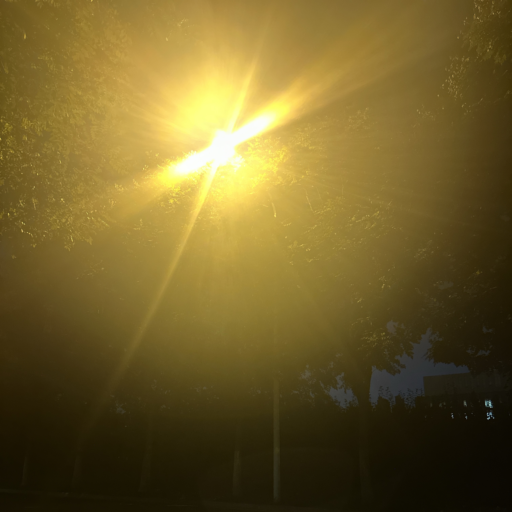} &
        \includegraphics[width=0.12\linewidth]{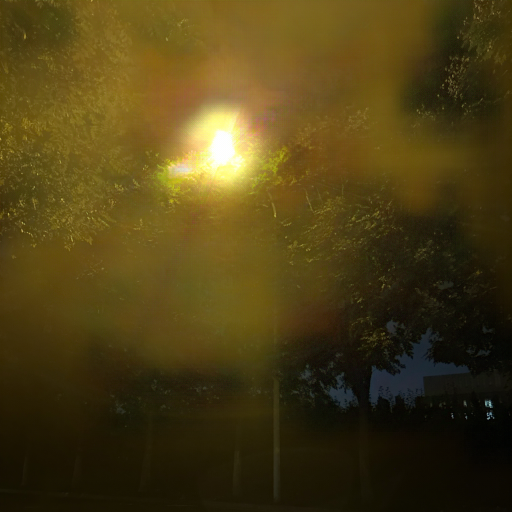} &
        \includegraphics[width=0.12\linewidth]{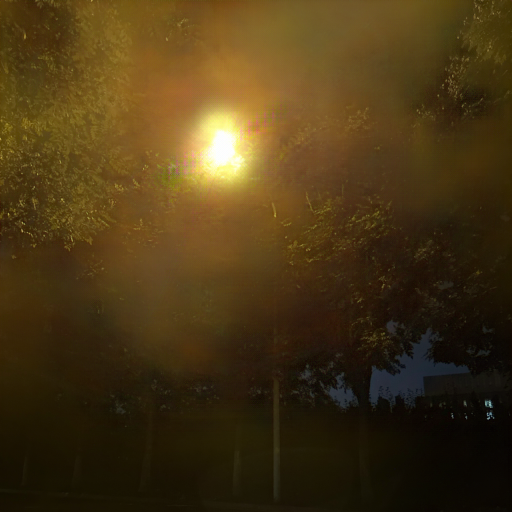} &
        \includegraphics[width=0.12\linewidth]{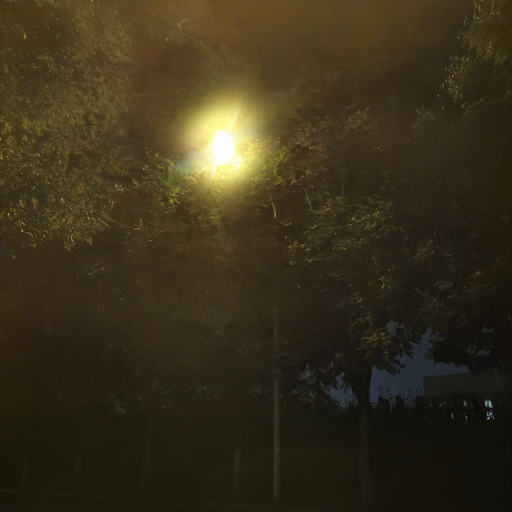} &
        \includegraphics[width=0.12\linewidth]{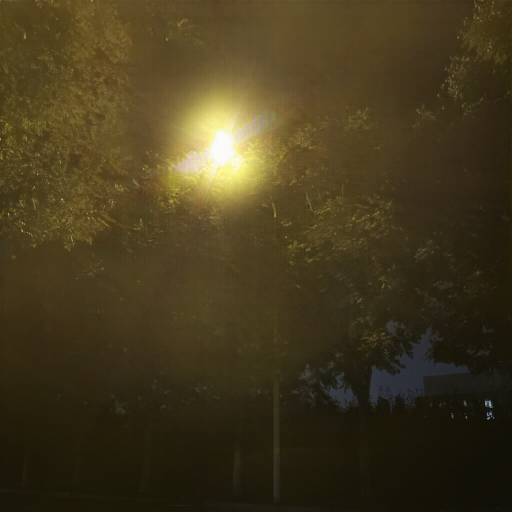} &
        \includegraphics[width=0.12\linewidth]{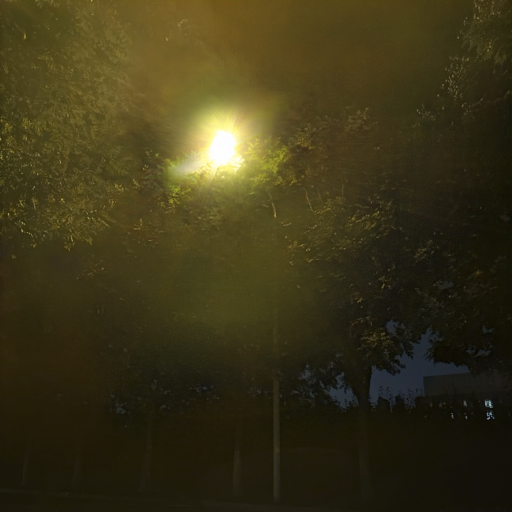} &
        \includegraphics[width=0.12\linewidth]{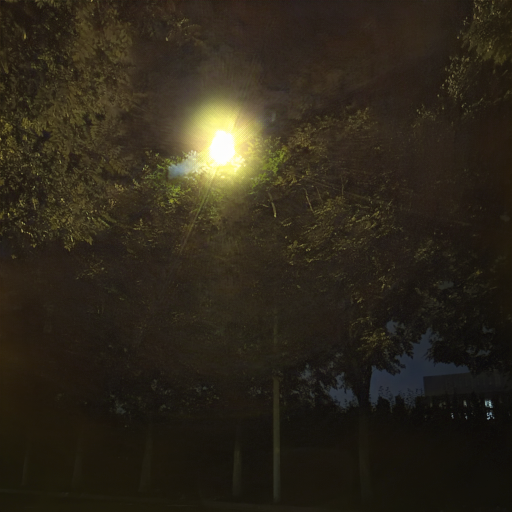} &
        \includegraphics[width=0.12\linewidth]{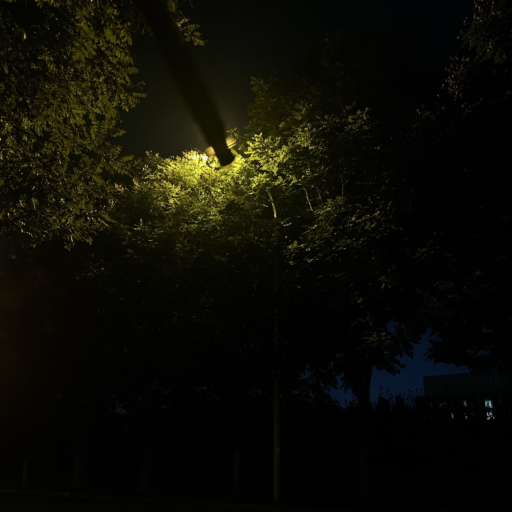} \\

        \includegraphics[width=0.12\linewidth]{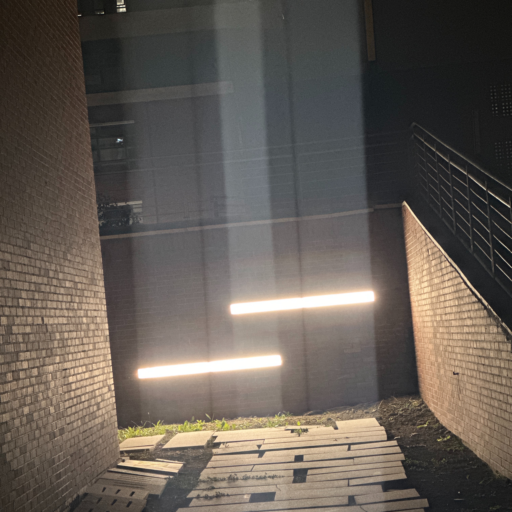} &
        \includegraphics[width=0.12\linewidth]{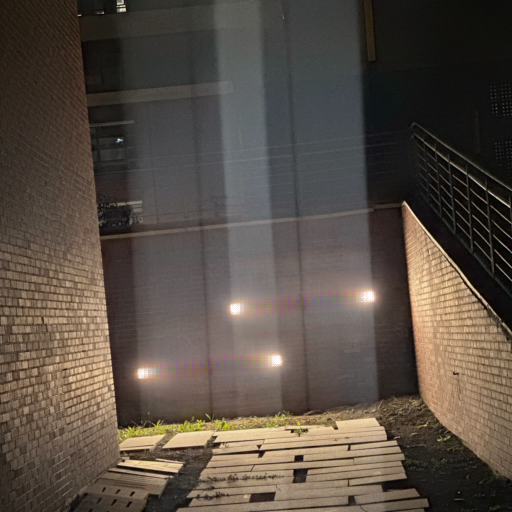} &
        \includegraphics[width=0.12\linewidth]{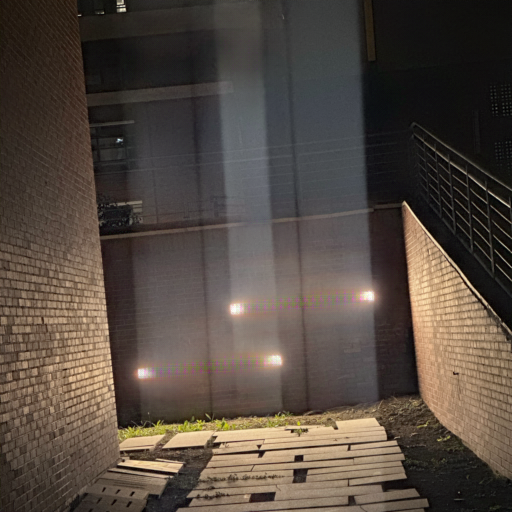} &
        \includegraphics[width=0.12\linewidth]{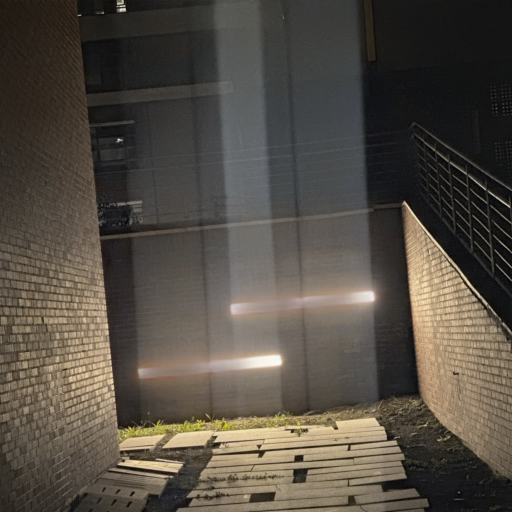} &
        \includegraphics[width=0.12\linewidth]{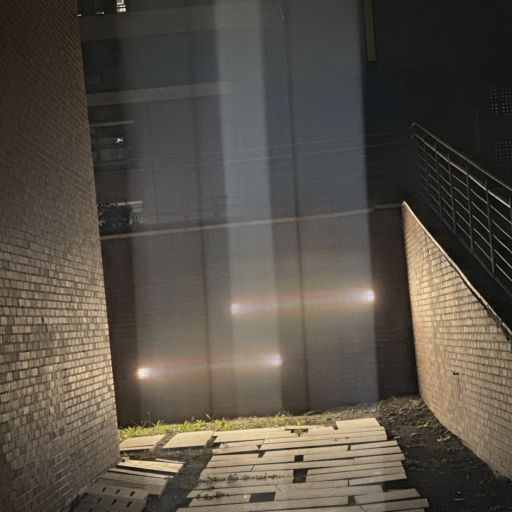} &
        \includegraphics[width=0.12\linewidth]{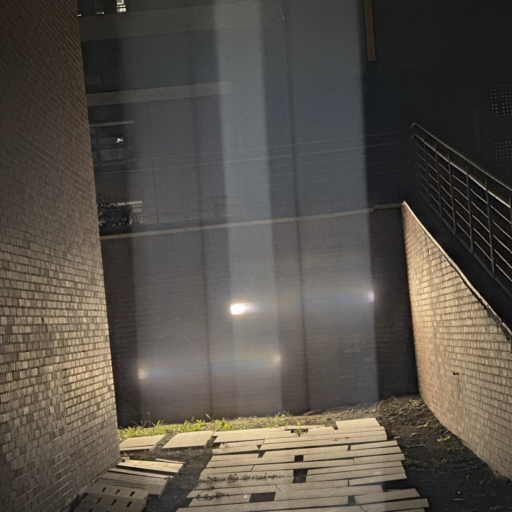} &
        \includegraphics[width=0.12\linewidth]{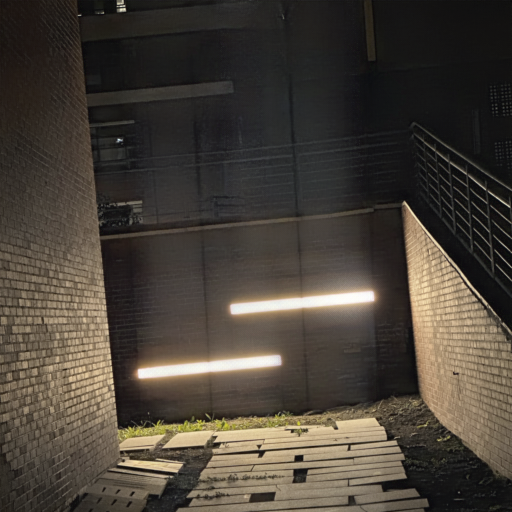} &
        \includegraphics[width=0.12\linewidth]{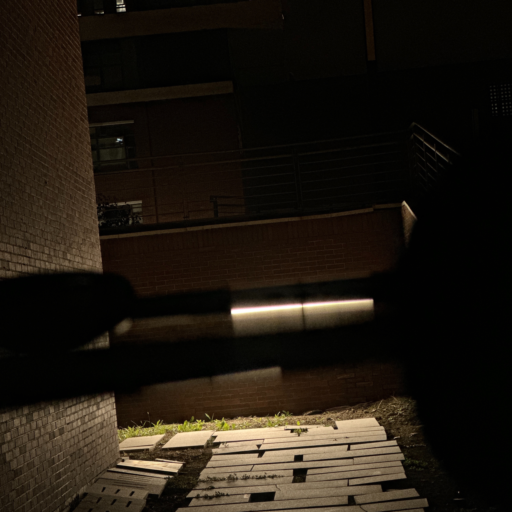} \\[-1mm]

        \makebox[0pt][c]{\footnotesize Input} & 
        \makebox[0pt][c]{\footnotesize Flare7K++\cite{DaiY2024TPAMI}} & 
        \makebox[0pt][c]{\footnotesize Kotp\cite{Kotp2024ICASSP}} & 
        \makebox[0pt][c]{\footnotesize DeflareMamba\cite{HuangY2025ACMMM}} & 
        \makebox[0pt][c]{\footnotesize FGRNet\cite{JieZ2025ICCV}} & 
        \makebox[0pt][c]{\footnotesize SGSFT\cite{MaT2025TASE}} & 
        \makebox[0pt][c]{\footnotesize \textbf{DeflareMambav2}} & 
        \makebox[0pt][c]{\footnotesize GT} \\

    \end{tabular}
    
    \vspace{-1mm} 

\caption{Visual comparisons on challenging real-world scenes. The top three rows are Flare7K-real, and the bottom three rows are FlareX.  Our DeflareMambav2 successfully removes complex lens flares while maintaining high background fidelity. From left to right: Input, Flare7K++ \cite{DaiY2024TPAMI}, Kotp \& Torki \cite{Kotp2024ICASSP}, DeflareMamba \cite{HuangY2025ACMMM}, FGRNet \cite{JieZ2025ICCV}, SGSFT \cite{MaT2025TASE}, \textbf{DeflareMambav2 (Ours)}, and Ground Truth.The first three rows show the results of Flare7k-real, and the last three rows show the results of FlareX.}

    \label{fig:compare}
\end{figure*}


As reported in Table \ref{tab:sota_comparison}, DeflareMambav2 achieves the highest Clean-PSNR and Clean-SSIM on the FlareX~\cite{QuL2025NeurIPS} test set, which rigorously reflects authentic restoration capability, while maintaining competitive Light-PSNR on the Flare7K-real dataset~\cite{DaiY2022NeurIPS}. As previously discussed, the ground truth of Flare7K-real contains residual flares, and this inherent impurity introduces an evaluation bias that accounts for the metric variations of our model on this dataset. Notably, the strong baseline established by DeflareMamba further validates the intrinsic potential of Mamba architectures for flare removal.


Fig.~\ref{fig:compare} presents qualitative visual comparisons illustrating that DeflareMambav2 achieves superior flare suppression while substantially outperforming competing baselines in preserving fine scene details. When handling streak light sources, spatially uniform approaches face a fundamental dilemma where they either mistakenly degrade the authentic source by identifying it as contamination or sacrifice restoration efficacy to maintain source integrity. Our model resolves this conflict through HB and the RSE. By breaking the constraints of uniform spatial operations via targeted heterogeneous processing, DeflareMambav2 maximizes flare removal capability while robustly preserving original luminous sources.

Finally, guided by prior learning, DeflareMambav2 proves to be highly parameter-efficient. Even when inclusive of the 1.65M-parameter FPN, the total model size remains merely 9.96M. This substantially undercuts the parameter overhead of current SOTA frameworks, achieving a remarkably lightweight architecture without compromising performance.

\subsection{Ablation Study}
\label{sec:ablation}

In this section, we conduct ablation studies to validate our network architecture and core components across three aspects. First, we evaluate the individual contributions of the three prior-aware modules by incrementally integrating them into the baseline and training to full convergence on the Flare7K++ pipeline. Second, we assess structural robustness against prior extraction failures. Third, we present a qualitative analysis of the FPN representations under different training strategies to demonstrate the critical role of optimization choices in explicit prior learning.

\subsubsection{Effectiveness of Proposed Three Modules}
\label{sec:ablation_modules}
\begin{table}[t] 
    \centering
        \caption{Ablation Study of different components. C-PSNR/SSIM denote Clean-PSNR/SSIM. L-PSNR denote Light-PSNR.}
    \label{tab:ablation}

    \resizebox{\linewidth}{!}{
        \begin{tabular}{ccc c c ccc c}
            \toprule

            \multicolumn{3}{c}{Modules} & 
            \multicolumn{2}{c}{FlareX } & 
            \multicolumn{3}{c}{Flare7K-real } & 
            \multirow{2}{*}{Params (M)} \\

            \cmidrule(lr){1-3} \cmidrule(lr){4-5} \cmidrule(lr){6-8}

            Radial unfold & HB & RSE & 
            C-PSNR & C-SSIM & 
            L-PSNR & PSNR & SSIM & \\
            
            \midrule
               &            &            & 28.553 & 0.758 & 27.64 & 27.415 & 0.893 & 8.05 \\

            \checkmark &            &            & 28.683 & 0.755 & 27.75 & 27.621 & 0.892 & 9.70 \\

            \checkmark & \checkmark &            & 29.224 & 0.760 & 27.96 & 27.616 & 0.895 & 9.70 \\

            \checkmark & \checkmark & \checkmark & \textbf{30.099} & \textbf{0.779} & \textbf{28.04} & \textbf{27.861} & \textbf{0.900} & 9.96 \\
            
            \bottomrule
        \end{tabular}
    }
\end{table}

To systematically validate our architectural contributions, we progressively integrate the three proposed components into the MambaIRv2 baseline. The quantitative results on the FlareX and Flare7K Real-test datasets are summarized in Table~\ref{tab:ablation}. As reported in the first row, the vanilla baseline yields suboptimal performance, highlighting the inherent limitations of operating without explicit prior guidance and customized modules.

Integrating the Radial unfold module yields consistent metric improvements. The 1.65M parameter increment originates entirely from the auxiliary FPN, while the Radial unfold operation itself is a parameter-free spatial transformation. Inspired by the radial attenuation characteristics of lens flares, this module prioritizes less contaminated regions by encoding clean pixels into the SSM memory first, building a robust historical state that subsequently guides the restoration of heavily corrupted areas, while naturally relegating the light source to the sequence terminus.

Building upon this ordering, the parameter-free Heterogeneous Branch (HB) deliberately breaks spatial uniformity by masking and protecting the light source. Since Radial unfold positions the light source at the sequence terminus, bypassing it avoids disrupting the hidden state transitions of preceding time steps, allowing the SSM to concentrate solely on flare decoupling without accounting for light source regions.

While the synergistic processing dictated by Radial unfold and HB excels at suppressing global flare, the Radial State-space Equation (RSE) is further introduced to address streak flares through customized detail modeling. By explicitly modulating the state transition matrices ($\bar{B}$ and $C$) via structural priors, RSE selectively erases severe flare artifacts while intensifying the query mechanism within the memory matrix, effectively restoring the underlying textures in regions left blank or blurry after removal.

\subsubsection{Robustness against Prior Failures}
\label{sec:ablation_robustness}

In real-world deployments, auxiliary modules like the Flare Prior Network (FPN) may occasionally fail to extract valid priors. To mitigate this risk, our architecture inherently incorporates a degradation mechanism. Upon receiving an empty prior signal, the proposed modules seamlessly revert to their conventional baseline formulations, preventing catastrophic structural collapse.

To quantitatively evaluate this resilience alongside the genuine contributions of each module, we simulate extreme failure scenarios on the fully converged DeflareMambav2. Specifically, we deliberately feed \texttt{None} to targeted modules during inference, compelling them to gracefully degrade into their baseline counterparts. As detailed in Table~\ref{tab:robustness}, the network maintains stable performance despite the absence of specific prior guidance, which not only verifies the real impact of these individual modules but also highlights the network's remarkable structural robustness.
\begin{table}[t]
    \centering
\caption{Robustness evaluation simulating prior extraction failures.}
    \label{tab:robustness}
    \resizebox{\linewidth}{!}{
        \begin{tabular}{l cc ccc}
            \toprule
            \multirow{2}{*}{Inference Configuration} & 
            \multicolumn{2}{c}{FlareX } & 
            \multicolumn{3}{c}{Flare7K-real} \\
            \cmidrule(lr){2-3} \cmidrule(lr){4-6}
             & Clean-PSNR & Clean-SSIM & Light-PSNR & PSNR & SSIM \\
            \midrule
            
            w/o Radial unfold  & 30.034 & 0.778 & 27.8 & 27.792 & 0.899 \\
            w/o HB  & 30.029 & 0.778 & 27.95 & 27.763 & \textbf{0.900} \\
            w/o RSE  & 27.492 & 0.725 & 26.67 & 26.600 & 0.885 \\
            
            \midrule
            \textbf{DeflareMambav2} & \textbf{30.099} & \textbf{0.779} & \textbf{28.04} & \textbf{27.861} & \textbf{0.900} \\
            \bottomrule
        \end{tabular}
    }
\end{table}

As Table~\ref{tab:robustness} shows, masking priors for Radial unfold and HB causes only marginal performance drops, confirming that the parameters learned under FPN guidance are sufficiently robust to operate without it, with the priors serving as safe topological constraints rather than fragile dependencies. Masking the RSE prior induces a comparatively larger decline, as RSE relies on it to dynamically modulate the hidden states ($\bar{B}$ and $C$); without prior guidance, the network reverts to static recurrence. Nevertheless, the representations internalized during FPN-guided training enable the model to maintain stable outputs without complete failure, demonstrating that the learned parameters generalize effectively even under prior-free inference.

\begin{figure*}[t]
    \centering
    \begin{minipage}{0.12\linewidth}
        \centering
        \includegraphics[width=\linewidth]{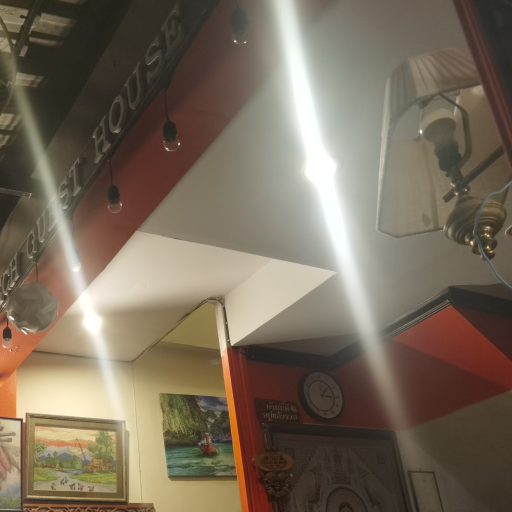}\vspace{2pt} \\
        \includegraphics[width=\linewidth]{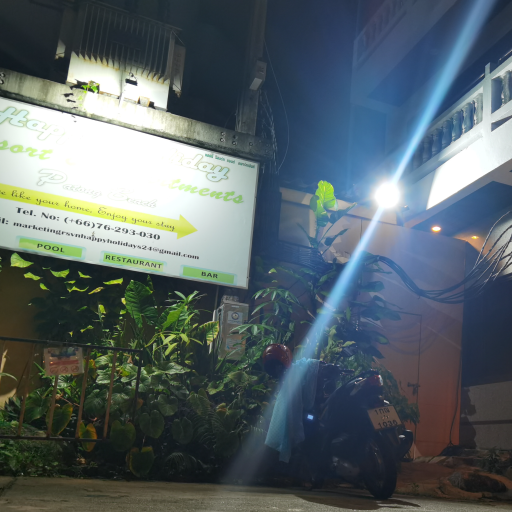}
        \centerline{\small (a) Input}
    \end{minipage}\hfill
    \begin{minipage}{0.12\linewidth}
        \centering
        \includegraphics[width=\linewidth]{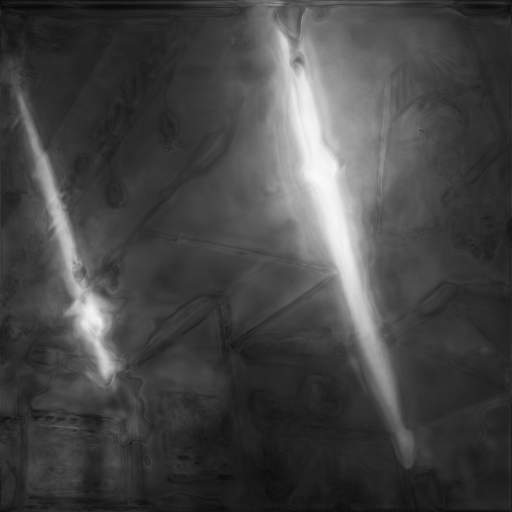}\vspace{2pt} \\
        \includegraphics[width=\linewidth]{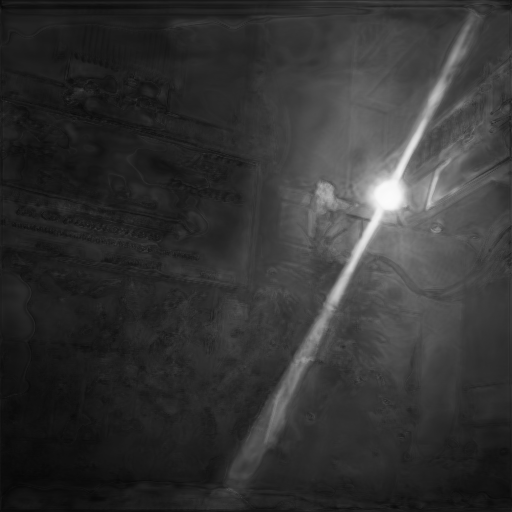}
        \centerline{\small (b) $P_{\text{flare}}$}
    \end{minipage}\hfill
    \begin{minipage}{0.12\linewidth}
        \centering
        \includegraphics[width=\linewidth]{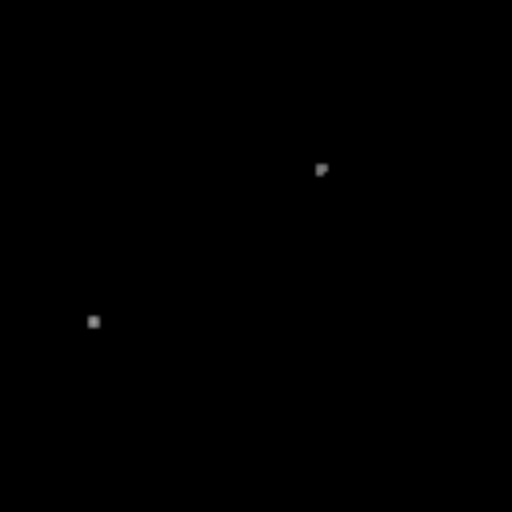}\vspace{2pt} \\
        \includegraphics[width=\linewidth]{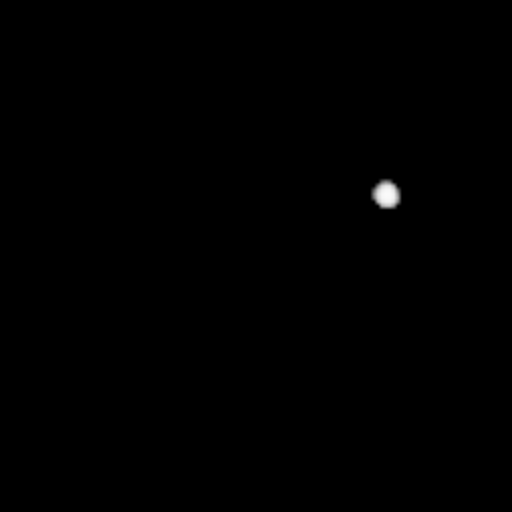}
        \centerline{\small (c) $P_{\text{mask}}$}
    \end{minipage}\hfill
    \begin{minipage}{0.12\linewidth}
        \centering
        \includegraphics[width=\linewidth]{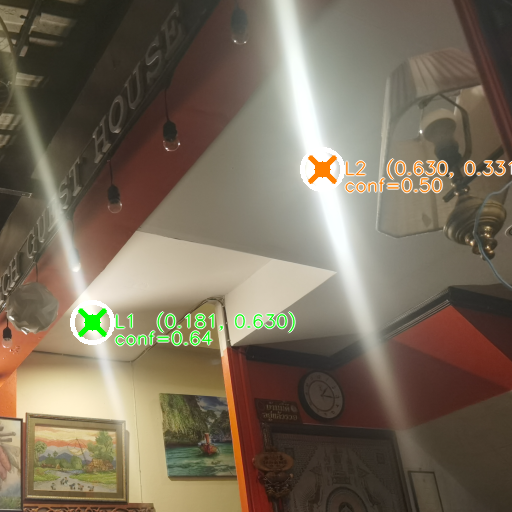}\vspace{2pt} \\
        \includegraphics[width=\linewidth]{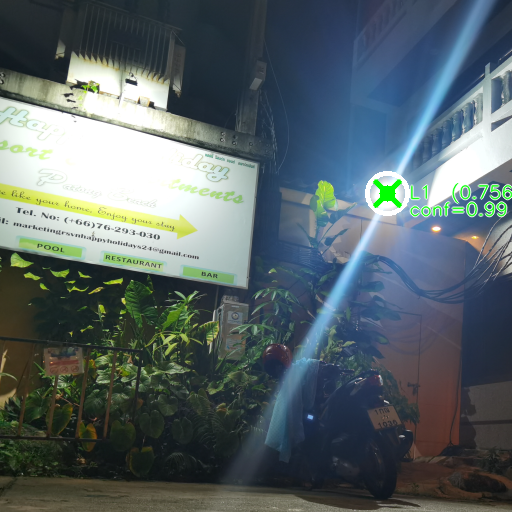}
        \centerline{\small (d) $P_{\text{position}}$}
    \end{minipage}\hfill
    \begin{minipage}{0.12\linewidth}
        \centering
        \includegraphics[width=\linewidth]{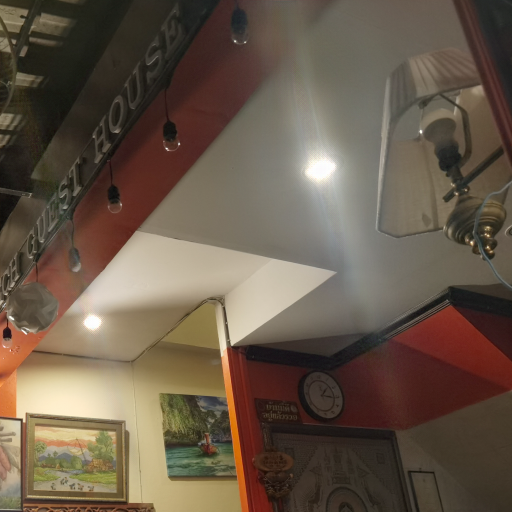}\vspace{2pt} \\
        \includegraphics[width=\linewidth]{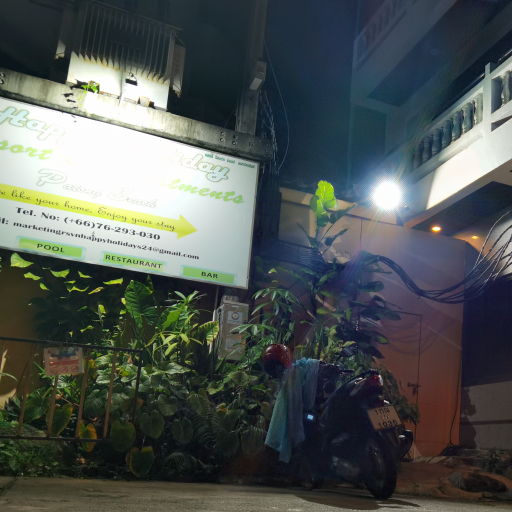} 
        \centerline{\small (e) w/o $P_{\text{flare}}$}
    \end{minipage}\hfill
    \begin{minipage}{0.12\linewidth}
        \centering
        \includegraphics[width=\linewidth]{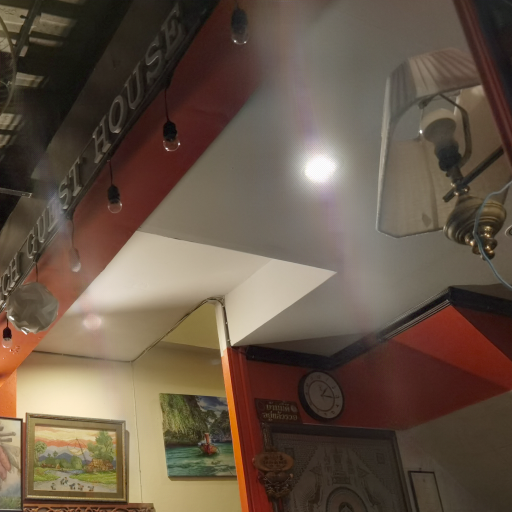}\vspace{2pt} \\
        \includegraphics[width=\linewidth]{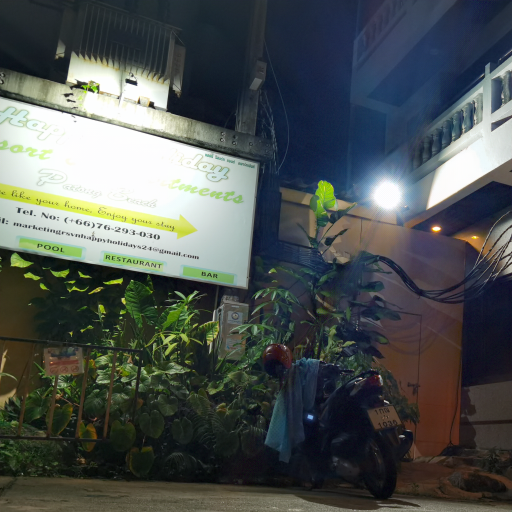}
        \centerline{\small (f) w/o $P_{\text{position}}$}
    \end{minipage}\hfill
    \begin{minipage}{0.12\linewidth}
        \centering
        \includegraphics[width=\linewidth]{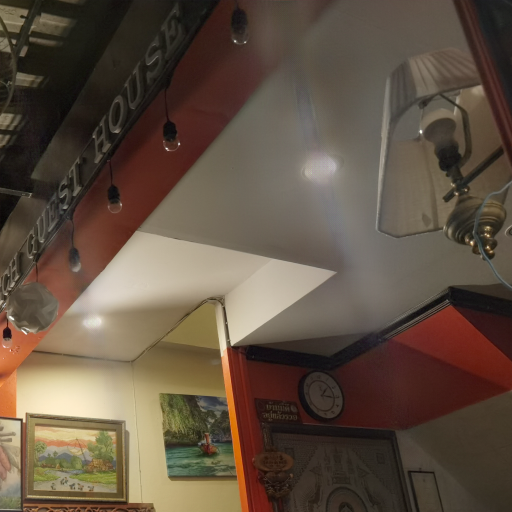}\vspace{2pt} \\
        \includegraphics[width=\linewidth]{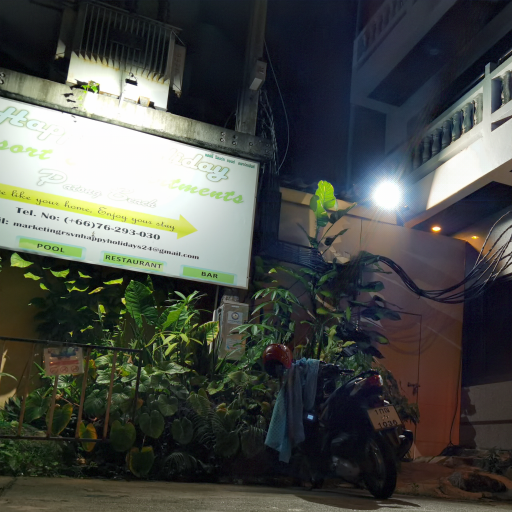}
        \centerline{\small (g) w/o $P_{\text{mask}}$}
    \end{minipage}\hfill
    \begin{minipage}{0.12\linewidth}
        \centering
        \includegraphics[width=\linewidth]{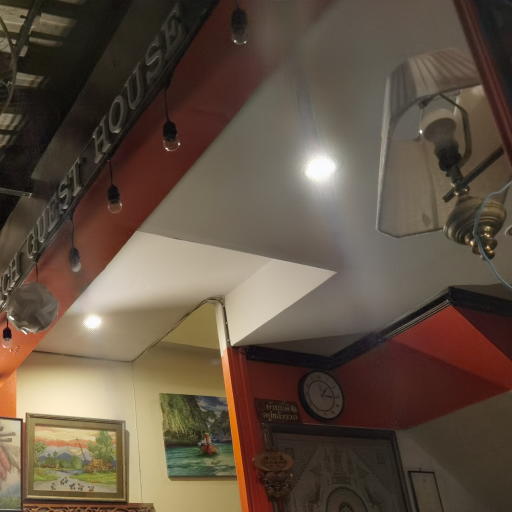}\vspace{2pt} \\
        \includegraphics[width=\linewidth]{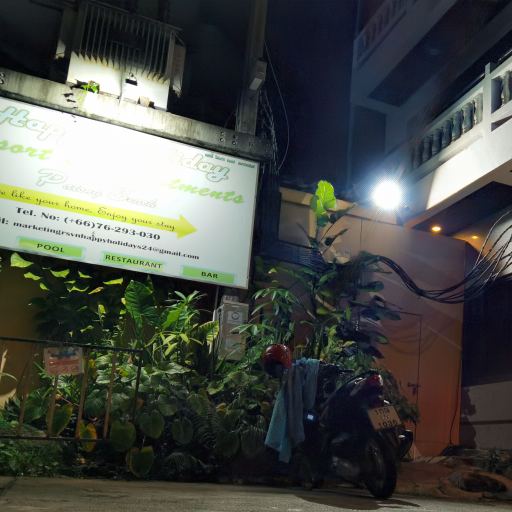}
        \centerline{\small (h) Ours}
    \end{minipage}
    
    \caption{Visualizations of the explicit priors generated by our FPN and the corresponding ablation results. From left to right: (a) Input images; the three explicit priors extracted by the FPN, including (b) flare priors ($P_{\text{flare}}$), (c) flare masks ($P_{\text{mask}}$), and (d) position priors ($P_{\text{position}}$); the restoration results of the main network when deprived of specific priors: (e) w/o $P_{\text{flare}}$, (f) w/o $P_{\text{position}}$, and (g) w/o $P_{\text{mask}}$; and (h) the final output of our complete DeflareMambav2 utilizing all priors.}
    \label{fpn_vis}
\end{figure*}

\begin{figure}[t]
    \centering

    \includegraphics[width=0.19\linewidth]{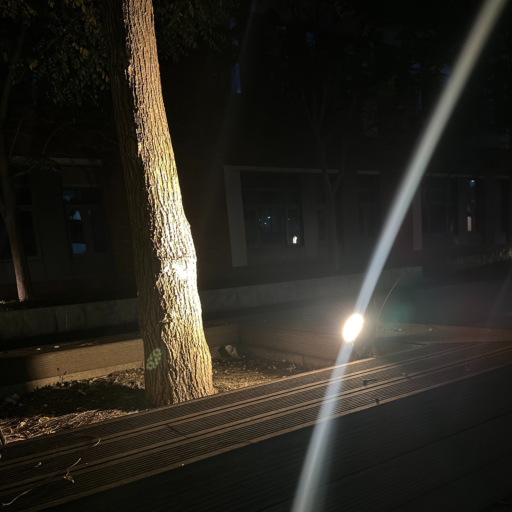} \hfill
    \includegraphics[width=0.19\linewidth]{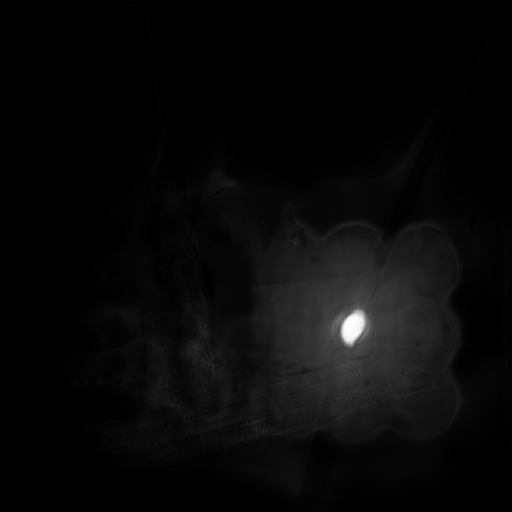} \hfill
    \includegraphics[width=0.19\linewidth]{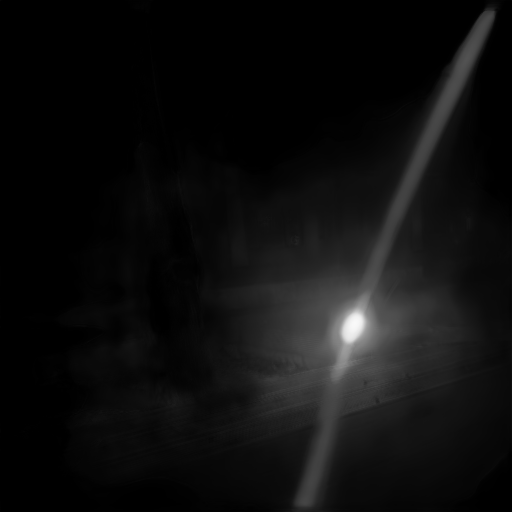} \hfill
    \includegraphics[width=0.19\linewidth]{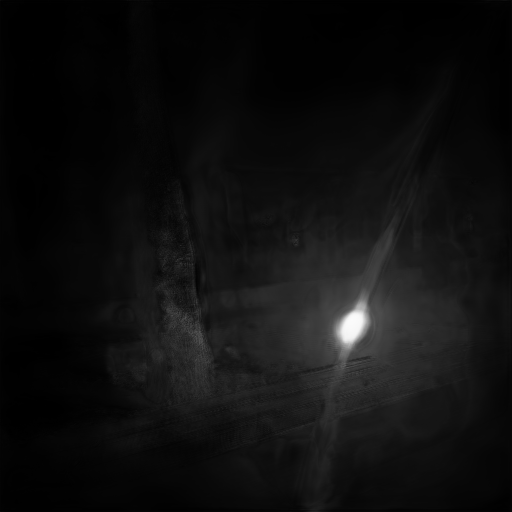} \hfill
    \includegraphics[width=0.19\linewidth]{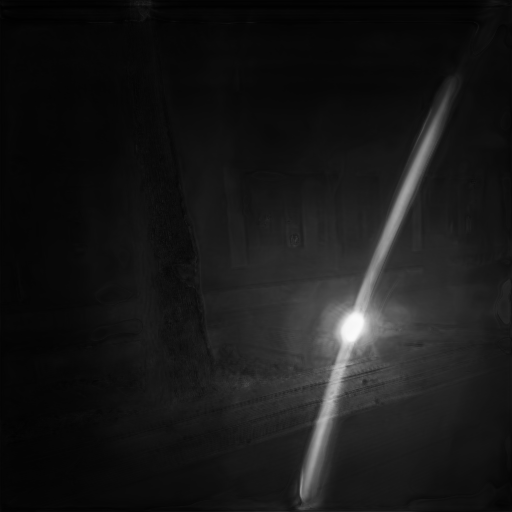} \\
    \vspace{1mm} 
    
    \includegraphics[width=0.19\linewidth]{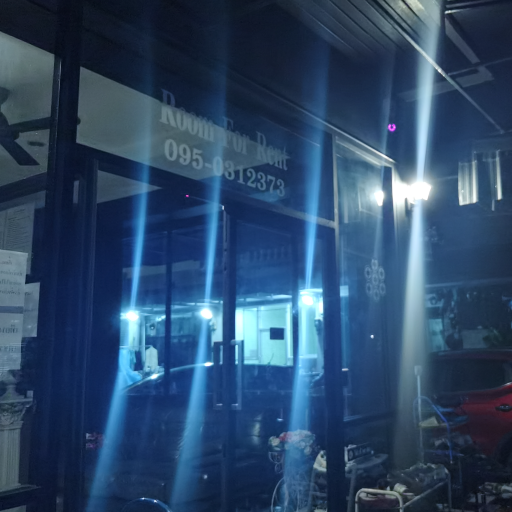} \hfill
    \includegraphics[width=0.19\linewidth]{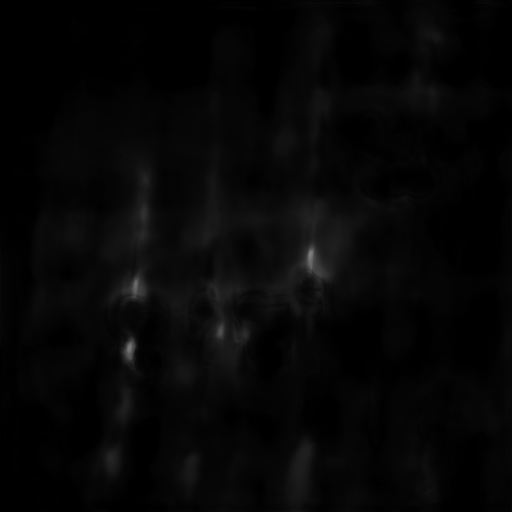} \hfill
    \includegraphics[width=0.19\linewidth]{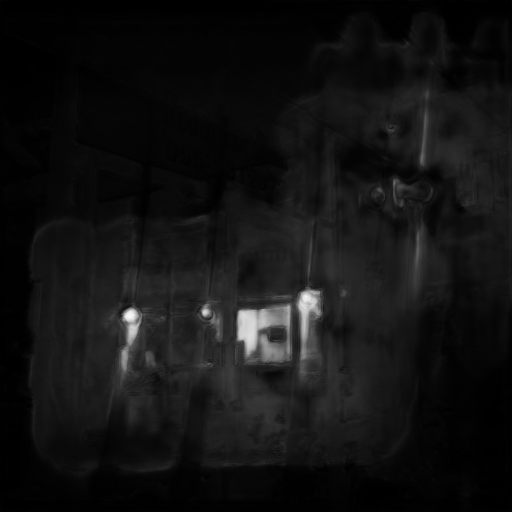} \hfill
    \includegraphics[width=0.19\linewidth]{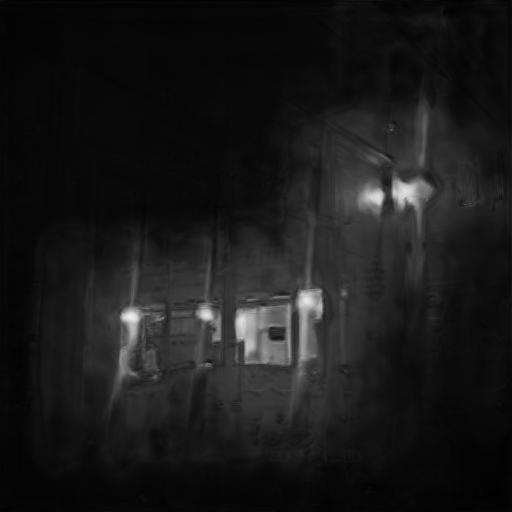} \hfill
    \includegraphics[width=0.19\linewidth]{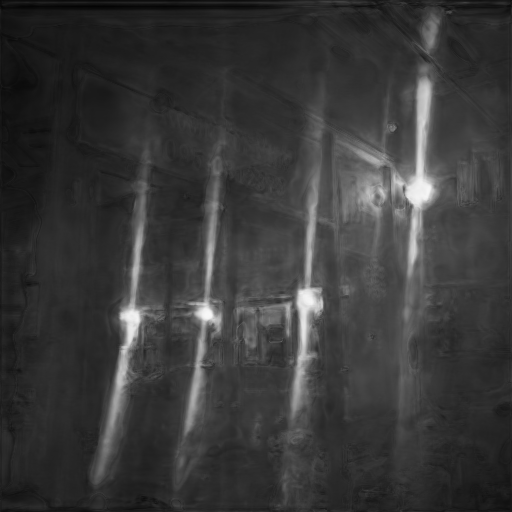} \\
    \vspace{1mm} 
    \makebox[0.19\linewidth]{\footnotesize (a) Input } \hfill
    \makebox[0.19\linewidth]{\footnotesize (b) Single 7K} \hfill
    \makebox[0.19\linewidth]{\footnotesize (c) Single 7K++} \hfill
    \makebox[0.19\linewidth]{\footnotesize (d) Multi 7K} \hfill
    \makebox[0.19\linewidth]{\footnotesize (e) Multi 7K++}
    
    \vspace{-2mm}
    \caption{Visual impact of different training strategies on prior extraction. Compared to baseline configurations, our optimal setup captures delicate streak structures without falsely activating on bright background regions. (a) Input. Models are trained using: (b) single-source synthetic flares of Flare7K, (c) single-source mixed flares of Flare7K++, (d) multi-source synthetic flares of Flare7K, and (e) multi-source mixed flares of Flare7K++ (Ours).}
    \label{fig:prior_comparison}
\end{figure}
As visually demonstrated in Fig.~\ref{fpn_vis}, our FPN exhibits strong capabilities. In multi-light scenarios, it precisely localizes individual light sources and generates segmentation masks ($P_{mask}$) and position priors ($P_{position}$). The FPN also shows robust resistance to bright background interferences, as evidenced by the low confidence assigned to the highly reflective whiteboard (second row, (b)), indicating that it does not rely solely on intensity for flare and source prediction.

\subsubsection{Impact of FPN Training Strategies}
\label{sec:ablation_fpn_strategy}

As illustrated in Figure~\ref{fig:prior_comparison}, we visualize the FPN outputs under various training data configurations. Since real flare templates in Flare7K++ are inherently off-center, the baseline strategy (b) is trained exclusively on single-source synthetic flares from Flare7K. Consequently, it lacks sensitivity to global intensity variations and misses faint flare artifacts. To leverage the real component, we align them to the center via spatial translation and zero-padding. Strategy (c), trained on a mixture of synthetic and real flares, successfully captures continuous contamination intensities. Finally, to further enhance robustness, we synthesize multi-source training samples. Strategy (d), trained on multi-source synthetic flares, forces the network to learn relative intensity differences through light source superposition. This effectively prevents over-saturated predictions, distinctly outperforming the single-source variant (b). Finally, (e) represents our optimal strategy leveraging multi-source mixed flares. It demonstrates superior capability in accurately estimating the pixel-wise contamination severity for both delicate streak structures and broad glare regions.

Furthermore, we conduct a quantitative ablation study by evaluating these FPN strategies while keeping the main network weights fixed. As reported in Table~\ref{tab:fpn_ablation}, our optimal configuration consistently achieves the highest performance across all metrics on both FlareX and Flare7K-real datasets, demonstrating the robustness and accuracy of our strategy.
\begin{table}[t]
    \centering
    \caption{Ablation study on FPN training strategies with a fixed main network. Configurations (b) through (e) correspond to the models visualized in Figure~\ref{fig:prior_comparison}.}
    \label{tab:fpn_ablation}
    \resizebox{\columnwidth}{!}{
        \begin{tabular}{l ccccc}
            \toprule
            \multirow{2}{*}{Configuration} & \multicolumn{2}{c}{FlareX} & \multicolumn{3}{c}{Flare7K-real} \\
            \cmidrule(lr){2-3} \cmidrule(lr){4-6}
             & Clean-PSNR & Clean-SSIM & Light-PSNR & PSNR & SSIM \\
            \midrule
            (b) Single 7K & 29.374 & 0.766 & 27.12 & 26.954 & 0.894 \\
            (c) Single 7K++ & 29.958 & 0.776 & 28.00 & 27.815 & 0.897 \\
            (d) Multi 7K & 29.944 & 0.777 & 27.95 & 27.775 & \textbf{0.900} \\
            \midrule
            (e) \textbf{Ours} & \textbf{30.099} & \textbf{0.779} & \textbf{28.04} & \textbf{27.861} & \textbf{0.900} \\
            \bottomrule
        \end{tabular}
    }
\end{table}

\section{Conclusion and Limitation}

In this paper, we present DeflareMambav2, a novel lens flare removal framework built upon spatial heterogeneity. Starting from the inherent duality between flare removal and light source preservation, we identify the fundamental limitations of spatially uniform processing and establish the necessity of heterogeneous treatment. To this end, we introduce a FPN to extract spatial priors and design three complementary modules for heterogeneous processing. Moreover, multi-source synthetic samples strengthen FPN robustness, and tailored losses yield more precise guidance for the main network. Extensive experiments demonstrate that DeflareMambav2 achieves state-of-the-art performance with the fewest parameters. 

\textbf{Limitation.} While the proposed light-source-guided Radial unfold strategy is highly effective for typical scenes, dense multi-light scenarios present a specific challenge. The complex superposition of multiple flares often introduces highly non-uniform intensity degradation, which may occasionally limit the robustness of the restoration. Additionally, the predominantly circular light sources in current synthesis pipelines lack geometric diversity, marginally restricting mask prediction and light source preservation. 
\bibliographystyle{IEEEtran} 
\bibliography{main} 
\end{document}